\documentclass[10pt,conference,a4paper]{IEEEtran}
\pdfoutput=1
\usepackage[pdftex]{graphicx}
\usepackage{amsmath,amssymb}
\usepackage{tikz}
\usepackage{pgfplots}
\usepackage{xspace}
\usepackage{tabularx}
\pgfplotsset{compat=1.13}
\usepackage{url}
\usepackage{hyperref}

\usepackage{float}

\makeatletter
\DeclareRobustCommand\onedot{\futurelet\@let@token\@onedot}
\def\@onedot{\ifx\@let@token.\else.\null\fi\xspace}

\def\etal{\emph{et al}\onedot}
\makeatother
\IEEEoverridecommandlockouts
\begin{document}
\title{Improving Multi-View Stereo\\via Super-Resolution}

\author{\IEEEauthorblockN{Eugenio Lomurno}
\IEEEauthorblockA{Politecnico di Milano, Italy\\
Email: eugenio.lomurno@polimi.it}
\and
\IEEEauthorblockN{Andrea Romanoni}
\IEEEauthorblockA{Politecnico di Milano, Italy\\
Email: andrea.romanoni@polimi.it*}
\thanks{*Work done prior to Amazon involvement of the author and does not reflect views of the Amazon company.}
\and
\IEEEauthorblockN{Matteo Matteucci}
\IEEEauthorblockA{Politecnico di Milano, Italy\\
Email: matteo.matteucci@polimi.it}}

\maketitle

\begin{abstract}
Today, Multi-View Stereo techniques are able to reconstruct robust and detailed 3D models, especially when starting from high-resolution images.
However, there are cases in which the resolution of input images is relatively low, for instance, when dealing with old photos, or when hardware constrains the amount of data that can be acquired.
In this paper, we investigate if, how, and how much increasing the resolution of such input images through Super-Resolution techniques reflects in quality improvements of the reconstructed 3D models, despite the artifacts that sometimes this may generate.
We show that applying a Super-Resolution step before recovering the depth maps in most cases leads to a better 3D model both in the case of PatchMatch-based and deep-learning-based algorithms.
The use of Super-Resolution improves especially the completeness of reconstructed models and turns out to be particularly effective in the case of textured scenes.
\end{abstract}

\IEEEpeerreviewmaketitle

\section{Introduction\label{intro}}
Recovering the 3D model of a scene captured by images is a relevant problem in a wide variety of scenarios, e.g., city mapping, archaeological heritage preservation, autonomous driving, and robot localization.
In the Computer Vision community, this task goes under the name of Multi-View Stereo (MVS), and it aims to reconstruct 3D models as accurately and completely as possible.

Currently, the most successful workflow to perform such reconstructions is the following. 
First, a Structure from Motion algorithm estimates camera parameters such as their positions and orientations \cite{schonberger2016structure}. 
Then, it follows the depth maps estimation step, for which the most common approaches rely on PatchMatch techniques \cite{bleyer2011patchmatch}, while recent learning-based algorithms have shown promising results \cite{yao2018mvsnet}. 
The former approaches lead to very accurate results, while the latter produce more complete models, even if they still suffer scalability issues nowadays. 
As last step, depth maps are projected on 3D space and fused together obtaining a dense point cloud.

Under controlled scenarios, in which the hardware adopted to collect the images is not subject to particular constraints, it is relatively easy to acquire high-resolution images and obtain a high-quality reconstruction of the scene following the previous pipeline.
However, in several cases, the input of an MVS method consists of low-resolution images. For instance, when power consumption constrains the hardware, e.g., with drones or telescopes, or when processing images taken in low-resolution such as with old photos.

In these cases, the recovered 3D model most likely lacks details or is incomplete, regardless of the adopted MVS algorithm.
We claim that by increasing artificially input images resolution it is possible to overcome this issue enhancing their information content and quality.
This is possible via different techniques which go under the name of Super-Resolution which, despite the possibility to generate some artifacts, have recently reached impressive performance in many application fields.

In this paper, we investigate the impact of upscaling through bicubic and deep-learning-based Single-Image Super-Resolution (SISR) low-resolution images for 3D Reconstruction with the aim of providing a general new step for MVS pipelines. 
In particular, we test SISR contribution over different MVS algorithms \cite{schoenberger2016mvs,romanoni2019tapa,gu2019cascade} applied to a wide set of datasets \cite{knapitsch2017tanks,schops2017multi}.
In many cases, SISR improves the reconstruction results, especially when dealing with outdoor and detailed scenes, obtaining remarkable improvements while dealing with very low-resolution starting images. 

\section{Related Work\label{related}}
In the literature, some works have already exploited Super-Resolution (SR) to improve the quality of 3D models from different perspectives. 
Goldl{\"u}cke \etal \cite{goldlucke2014super} proposed a variational method to improve 3D models appearance by estimating SR textures. 
Another model-based SR method was proposed by Maier \etal \cite{maier2015super}, who fused multiple RGB-D and colour images into SR depth and RGB keyframes to enhance the texture mapping process. 
More recently, Li \etal \cite{li20193d} proposed a novel model-based SR method which better exploits geometric features to enhance the texture applied to a 3D model.

Other approaches exploiting SR in the 3D reconstruction realm aims to increase depth maps resolution.
Lei \etal \cite{lei2017depth} relied on bilinear interpolation of multiple depth maps to increase the resolution of a single depth map. 
While, the authors in \cite{zuo2019multi} and \cite{voynov2019perceptual} have used high-resolution RGB images to guide a DNN to increase depth maps resolution.

Differently from previous works, we look at improving models geometry instead of their texture appearance, by applying SR directly on input images. 
We argue that SR can improve the reconstruction from low-resolution images, and different stages of a 3D reconstruction pipeline could benefit from the availability of SR images, e.g., camera calibration and mesh refinement.
Surprisingly, to the best of our knowledge, no paper has ever analyzed if and to what extent MVS 3D reconstruction pipelines can benefit from input images enhanced trough SR. 

\subsection{Single-Image Super-Resolution}
Single-Image Super-Resolution (SISR) aims at recovering a high-resolution image from a single low-resolution image. 
In the last few years, we have seen how modern Deep Learning pipelines overtook non-learning-based algorithms, as nearest-neighbours and bicubic interpolation. 

As first attempt, Dong \etal \cite{dong2015image} proposed a CNN to learn how to map low- to high-resolution images. 
This network architecture has been extended with a combination of new layers and skip-connections by Kim \etal \cite{kim2016accurate}. 
Subsequently, other methods have exploited different combinations of residual and dense connections \cite{lim2017enhanced,zhang2018residual}.
Recent works show that networks with novel feedback mechanisms further improve the quality of the SR images. 
For instance, Haris \etal \cite{haris2018deep} use Back-Projection to provide error feedback during the learning process, while, Li \etal \cite{li2019feedback} combine a feedback block with curriculum learning.
For a more detailed review of SISR techniques, we refer the reader to \cite{yang2019deep}.

\subsection{Multi-View Stereo}
MVS aims at recovering a dense 3D representation of a scene perceived by a set of calibrated cameras.
The main step adopted by the most successful MVS methods is depth maps estimation, i.e., the process of computing the depth of each pixel belonging to each image. 
Once computed, these maps are fused into a dense point cloud, or a volumetric representation.

Most performing depth estimation approaches are based on the PatchMatch algorithm \cite{bleyer2011patchmatch}, which relies on the idea of choosing for each pixel a random guess of the depth and then propagate the most likely estimates to its neighbourhood.
The work proposed by Sch\"{o}nberger \etal \cite{schonberger2016pixelwise}, named COLMAP, can be considered the cornerstone of modern PatchMatch-based algorithms: it is a robust framework able to process high-quality images and to jointly estimate pixel-wise camera visibility, as well depth and normal maps for each view. 
Since this method heavily relies on the Bilateral NCC Photometric-Consistency, it often fails in recovering areas with low texture.
Recently, to compensate for this, TAPA-MVS \cite{romanoni2019tapa} proposed to explicitly handle textureless regions by propagating in a planar-wise fashion the valid depth estimates to neighbouring textureless areas. 
Kuhn \etal \cite{kuhn2019plane} extended this method with a hierarchical approach improving the robustness of the estimation process.

Another family of MVS algorithms relies on Deep Learning. 
DeepMVS \cite{huang2018deepmvs} and MVSNet \cite{yao2018mvsnet} were the first approaches proposing an effective MVS pipeline based on DNNs. 
For each camera, they build a cost volume by projecting nearby images on planes at different depths, then they classify \cite{huang2018deepmvs} or regress \cite{yao2018mvsnet} the best depth for each pixel.
Yao \etal \cite{yao2019recurrent} introduced an RNN to regularize the cost volume, while Luo \etal \cite{luo2019p}  built a model to learn how to aggregate the cost to compute a more robust depth estimate. MVS-CRF \cite{xue2019mvscrf}, finally, refines the MVSNet estimate through Markov Random Field, and Point-MVSNet \cite{chen2019point} through a graph-based neural architecture.
The huge limitation of deep-learning-based approaches is their computational complexity.
Usually, they cannot handle high-resolution images as both memory and time costs grow cubically as the volume resolution increases, causing a limitation on accuracy and completeness of the reconstructed models.
The best attempt to handle this problem is the work of Xiaodong \etal \cite{gu2019cascade}, named CasMVSNet, in which they applied a coarse-to-fine approach that considerably improves the scalability of MVSNet-based methods.

\section{Improving Multi-View Stereo via SR\label{pipeline}}
In the following, we define the concise notation that we use in the rest of the paper to avoid ambiguity.
Let $\mathcal{S}$ be a set of $N$ image sequences $\mathcal{I}_{n}$, each one composed by images from the same scene captured by one or more cameras, and $\mathcal{SR}_{k}$ a Single-Image Super-Resolution (SISR) function with scale factor $k$.
For each sequence of images $\mathcal{I}_{n}$ with dimensions $w_{n}$ and $h_{n}$, we compute the Super-Resolution (SR) image set $\mathcal{I}_{n}^{SR,k} = \mathcal{SR}_k(\mathcal{I}_{n})$ composed by images with width $kw_n$ and height $kh_n$. We call $\mathcal{S}^{SR,k}$ the set of these new sequences.
We compute camera parameters $\mathcal{C}_{n}^{k}$ for the new set used to compute the sparse point cloud, and then to undistort the SR sequence, obtaining the sequence of undistorted images $\mathcal{U}_{n}^{SR,k}$.
Let $\mathcal{MVS}$ be a generic Multi-View Stereo (MVS) pipeline able to process high-resolution images. We define the set of SR depth maps $\mathcal{M}_{n}^{SR,k} = \mathcal{MVS}(\mathcal{U}_{n}^{SR,k})$ and, if the algorithm requires it, we can filter the depth maps before fusing them to obtain the dense point cloud.

\subsection{Interpolation-based and learning-based SISR}
To provide a detailed overview of SISR impact in 3D reconstruction field, in our experiments we have chosen to improve images resolution with both the well known bicubic interpolation algorithm and the Deep Back-Projection Network (DBPN) \cite{DBPN2019}, a deep-learning-based SISR model made by Haris \etal.

The DBPN architecture is composed by multiple iterative up- and down- layers grouped into units and used to provide a projection error feedback mechanism, leading to numerous degraded and high-resolution hypothesis images that the network uses to improve the output result.
In the last revision, the authors have implemented dense connections, adversarial loss and recurrent layers, making the entire architecture more scalable and performing.

To keep stable computational efforts and minimize the artifacts presence, which is more frequent for high enlargement factors, we set the scale factor $k=2$ for both the algorithms.
For DBPN we have used the ``DBPN-RES-MR64-3" model provided by the authors which is, according to them, the most performing among the others for the chosen scale factor.



\subsection{PatchMatch-based and learning-based MVS}
We are interested in evaluating the contribution of SISR on top of MVS pipelines to find correlations between the enhanced input images and the obtained 3D models.
We tested the proposed approach over PatchMatch-based and deep-learning-based MVS pipelines, which actually represent the state-of-the-art in the 3D reconstruction field.

In this paper, we focus on the most commonly used PatchMatch method, which is COLMAP \cite{schonberger2016pixelwise}. 
Besides, we test a more recent PatchMatch approach TAPA-MVS \cite{romanoni2019tapa} that explicitly deals with textureless regions.
In fact, COLMAP MVS step is heavily based on the Bilateral Normalized Cross-Correlation (NCC) Photometric-Consistency and thus tends to produce artifacts and poor estimates in textureless regions such as monochromatic and reflective surfaces.
To address this problem, we run experiments with COLMAP default parameters setup but we also modify its PatchMatch parameters for textureless datasets to increase the robustness of the depth estimate trading-off with a higher computational cost.

\begin{figure*}[tbp]
    \centering
    \setlength{\tabcolsep}{-2.5pt}
    \begin{tabular}{cccc}
    \includegraphics[width=0.25\textwidth,height=0.15\textwidth]{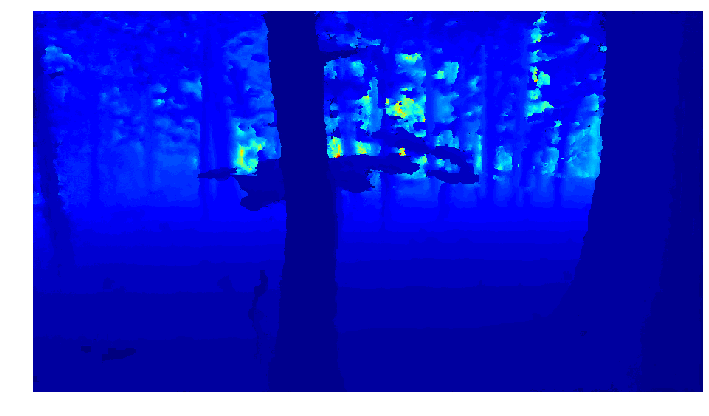}&
    \includegraphics[width=0.25\textwidth,height=0.15\textwidth]{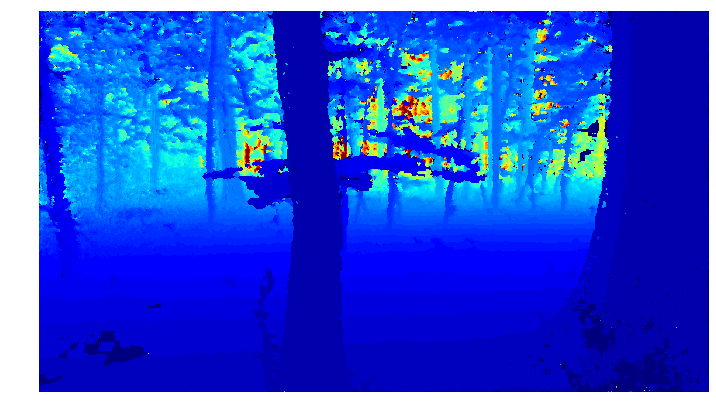} \: &
    \includegraphics[width=0.25\textwidth,height=0.15\textwidth]{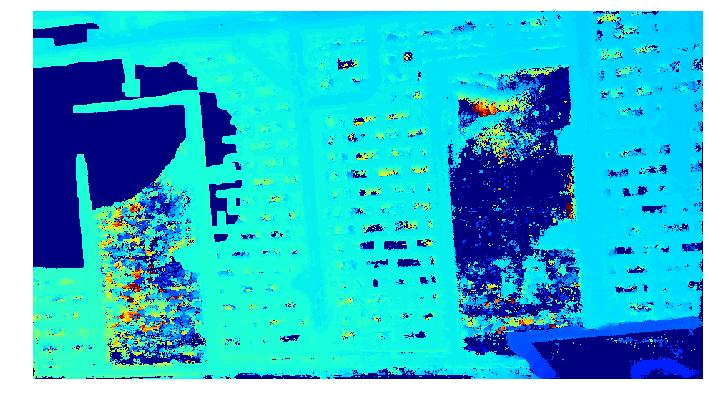}&
    \includegraphics[width=0.25\textwidth,height=0.15\textwidth]{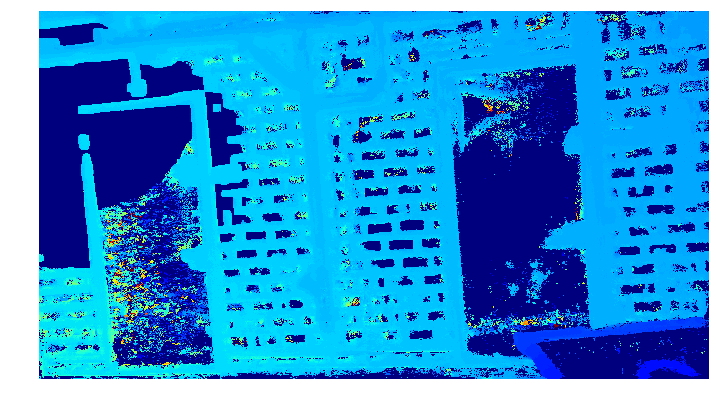}\vspace{-5pt}\\
    \multicolumn{2}{c}{(a)} & \multicolumn{2}{c}{\:\:(b)}\vspace{-2.5pt}\\
    \end{tabular}
    \caption{COLMAP disparity maps of a sample from forest dataset (a) and a sample from storage\_room\_2 one. 
    In both the cases, on the left the low-resolution estimation, on the right the DBPN SR one.
    In both the cases the disparity and the color maps ranges are kept constant between low- and DBPN SR pairs.}
    \label{fig:depthmapscolmap}
\end{figure*}

\begin{table*}[tbp]
    \caption{F1 scores on the ETH3D low-resolution multi-view train set with COLMAP. We compare scores starting from low-resolution images against the ones obtained from bicubic and Deep Back-Projection Network Super-Resolution}
    \label{tab:COLMAPtrain}
    \centering
    \setlength{\tabcolsep}{2.5px}
    \begin{tabular}{c|ccc|ccc|ccc|ccc|ccc}
    $\tau$& \multicolumn{3}{c}{Overall} & \multicolumn{3}{c}{Indoor} & \multicolumn{3}{c}{Outdoor}& \multicolumn{3}{c}{Textured} & \multicolumn{3}{c}{Textureless}\\
     (cm) & low-res & bicubic & DBPN & low-res & bicubic & DBPN & low-res & bicubic & DBPN & low-res & bicubic & DBPN & low-res & bicubic & DBPN\\
    1   & 35.80 & 39.85 & $\mathbf{40.00}$    & 40.68 & $\mathbf{43.68}$ & 43.42   & 32.55 & 37.29 & $\mathbf{37.71}$   & 32.07 & 36.84 & $\mathbf{37.28}$   & 38.29 & $\mathbf{41.85}$ & 41.80\\
    2   & 53.41 & 54.83 & $\mathbf{54.98}$    & 56.15 & $\mathbf{57.37}$ & 57.09   & 51.59 & 53.14 & $\mathbf{53.58}$   & 50.42 & 52.25 & $\mathbf{52.75}$   & 55.40 & $\mathbf{56.56}$ & 56.47\\
    5   & 72.16 & 72.58 & $\mathbf{72.64}$    & $\mathbf{74.35}$ & 74.05 & 73.62   & 70.70 & 71.59 & $\mathbf{71.99}$   & 69.67 & 70.96 & $\mathbf{71.43}$   & $\mathbf{73.82}$ & 73.65 & 73.45\\
    10   & 81.83 & $\mathbf{82.17}$ & 82.13    & $\mathbf{83.97}$ & 83.24 & 82.69   & 80.40 & 81.46 & $\mathbf{81.76}$   & 79.96 & 81.3 & $\mathbf{81.69}$   & $\mathbf{83.07}$ & 82.75 & 82.43\\
    20   & 88.98 & $\mathbf{89.20}$ & 89.14    & $\mathbf{91.50}$ & 90.55 & 90.11   & 87.30 & 88.30 & $\mathbf{88.50}$   & 87.36 & 88.41 & $\mathbf{88.69}$   & $\mathbf{90.06}$ & 89.73 & 89.44\\
    50   & 95.29 & $\mathbf{95.70}$ & $\mathbf{95.70}$    & $\mathbf{97.33}$ & 97.19 & 97.15   & 93.93 & 94.71 & $\mathbf{94.75}$   & 94.01 & 94.78 & $\mathbf{94.84}$   & 96.15 & $\mathbf{96.31}$ & 96.28\\
    \end{tabular}
\end{table*}

\begin{table*}[tbp]
    \caption{F1 scores on the ETH3D low-resolution multi-view train set with TAPA-MVS. We compare scores starting from low-resolution images against the ones obtained from bicubic and Deep Back-Projection Network Super-Resolution}
    \label{tab:TAPAtrain}
    \centering
    \setlength{\tabcolsep}{2.5px}
    \begin{tabular}{c|ccc|ccc|ccc|ccc|ccc}
    $\tau$  & \multicolumn{3}{c}{Overall} & \multicolumn{3}{c}{Indoor} & \multicolumn{3}{c}{Outdoor}& \multicolumn{3}{c}{Textured} & \multicolumn{3}{c}{Textureless}\\
    
    (cm)& low-res & bicubic & DBPN & low-res & bicubic & DBPN & low-res & bicubic & DBPN & low-res & bicubic & DBPN & low-res & bicubic & DBPN\\

    1   & 38.87 & $\mathbf{42.45}$ & 42.33    & 45.22 & $\mathbf{45.29}$ & 44.67   & 34.64 & 40.56 & $\mathbf{40.77}$   & 34.77 & 40.09 & $\mathbf{40.22}$   & 41.60 & $\mathbf{44.02}$ & 43.73\\
    2   & 55.12 & $\mathbf{56.22}$ & 56.08    & $\mathbf{58.20}$ & 57.30 & 56.78   & 53.07 & 55.51 & $\mathbf{55.62}$   & 51.21 & 54.33 & $\mathbf{54.38}$   & $\mathbf{57.73}$ & 57.49 & 57.22\\
    5   & 72.54 & $\mathbf{72.8}$ & 72.78    & $\mathbf{73.59}$ & 72.48 & 72.15   & 71.84 & 73.01 & $\mathbf{73.19}$   & 69.81 & 71.98 & $\mathbf{72.20}$   & $\mathbf{74.36}$ & 73.34 & 73.16\\
    10   & 81.65 & 82.28 & $\mathbf{82.41}$    & $\mathbf{82.34}$ & 82.10 & 81.94   & 81.19 & 82.41 & $\mathbf{82.72}$   & 80.38 & 82.23 & $\mathbf{82.62}$   & $\mathbf{82.49}$ & 82.31 & 82.27\\
    20   & 88.45 & 89.51 & $\mathbf{89.65}$    & 89.81 & 90.32 & $\mathbf{90.35}$   & 87.54 & 88.97 & $\mathbf{89.18}$   & 87.84 & 89.38 & $\mathbf{89.71}$   & 88.86 & 89.59 & $\mathbf{89.62}$\\
    50   & 95.10 & 96.06 & $\mathbf{96.13}$    & 96.40 & 97.56 & $\mathbf{97.59}$   & 94.23 & 95.06 & $\mathbf{95.16}$   & 95.15 & 95.69 & $\mathbf{95.89}$   & 95.07 & $\mathbf{96.3}$ & 96.29\\
    \end{tabular}
\end{table*}

\begin{table*}[tbp]
    \caption{F1 scores on the ETH3D low-resolution multi-view train set with CasMVSNet. We compare scores starting from low-resolution images against the ones obtained from bicubic and Deep Back-Projection Network Super-Resolution}
    \label{tab:CasMVSNettrain}
    \centering
    \setlength{\tabcolsep}{2.5px}
    \begin{tabular}{c|ccc|ccc|ccc|ccc|ccc}
    $\tau$ & \multicolumn{3}{c}{Overall} & \multicolumn{3}{c}{Indoor} & \multicolumn{3}{c}{Outdoor}& \multicolumn{3}{c}{Textured} & \multicolumn{3}{c}{Textureless}\\
    
    (cm) & low-res & bicubic & DBPN & low-res & bicubic & DBPN & low-res & bicubic & DBPN & low-res & bicubic & DBPN & low-res & bicubic & DBPN\\

    1   & 38.28 & 39.58 & $\mathbf{39.59}$    & 37.24 & 38.43 & $\mathbf{38.46}$   & 38.97 & 40.34 & $\mathbf{40.35}$   & 38.18 & 38.77 & $\mathbf{38.78}$   & 38.34 & 40.11 & $\mathbf{40.13}$\\
    2   & 49.00 & 49.65 & $\mathbf{49.66}$    & 48.08 & 48.07 & $\mathbf{48.09}$   & 49.61 & 50.70 & $\mathbf{50.71}$   & 47.87 & $\mathbf{48.42}$ & 48.41   & 49.75 & 50.47 & $\mathbf{50.49}$\\
    5   & 60.58 & $\mathbf{61.17}$ & 61.15    & $\mathbf{61.10}$ & 60.49 & 60.40   & 60.25 & 61.63 & $\mathbf{61.65}$   & 57.20 & 58.34 & $\mathbf{58.36}$   & 62.84 & $\mathbf{63.06}$ & 63.01\\
    10   & 67.59 & $\mathbf{68.43}$ & 68.37    & $\mathbf{69.39}$ & 69.25 & 69.10   & 66.39 & $\mathbf{67.89}$ & $\mathbf{67.89}$   & 62.66 & 63.92 & $\mathbf{63.93}$   & 70.87 & $\mathbf{71.44}$ & 71.33\\
    20   & 74.00 & $\mathbf{74.93}$ & 74.84    & 77.33 & $\mathbf{77.44}$ & 77.25   & 71.79 & $\mathbf{73.26}$ & 73.23   & 67.93 & $\mathbf{69.24}$ & 69.20   & 78.06 & $\mathbf{78.72}$ & 78.59\\
    50   & 82.60 & 83.59 & $\mathbf{83.61}$    & 87.14 & $\mathbf{87.31}$ & $\mathbf{87.31}$   & 79.56 & 81.10 & $\mathbf{81.14}$   & 75.98 & 77.69 & $\mathbf{77.79}$   & 87.00 & $\mathbf{87.52}$ & 87.49\\
    \end{tabular}
\end{table*}

Specifically, while reconstructing from an image set with a high percentage of textureless surfaces, we reduce the minimum NCC threshold and increase the window radius by the same scale factor chosen for SR, i.e., $k=2$.
Then we filter the resulting depth maps with a speckle filter algorithm before fusing them.
In detail, this filter family is based on two main parameters, i.e. the max depth range $d$ and the max speckle size $s$, which have been tuned to maximize average performance: we have fixed $d=0.5$ and $s=\frac{depth\_map\_area}{100}$ to keep constant the percentage of filtered area for each sequence in each resolution.

In our experiments we have adopted the aforementioned COLMAP setup for each set of images classified as textureless, while the normal pipeline with default parameters and without filters for the rest.
Regarding TAPA-MVS, we have adopted its default version for the entire set of experiments.

We also investigated the behaviour of SISR on top of a deep-learning-based MVS pipeline.
For this purpose, we used CasMVSNet, a Deep Learning architecture based on MVSNet with the addition of a new cost volume built upon a feature pyramid encoding geometry and context at gradually finer scales. 
It narrows the depth or disparity range for every stage thanks to a prediction made from the previous stage and then gradually increases the cost volume resolution to obtain accurate output.
We use a pre-trained model provided by the authors with the hyperparameters described in~\cite{gu2019cascade}.

Due to computational and pre-training constraints, we were forced to keep the maximum size of input images to the algorithm default value of 1152x864. 
In the experiments, however, we could still appreciate SR effects in the deep-learning-based context, keeping in mind the existence of a wide margin of improvement related to the maximum input dimensions.

\section{Experimental Results\label{experiments}}
All the experiments have been computed on an Intel(R) Xeon(R) CPU E5-2630 v4 @ 2.20GHz with an Nvidia GTX 1080Ti.
Several MVS benchmarks are publicly available, such as Tanks and Temples \cite{knapitsch2017tanks}, DTU-MVS \cite{jensen2014large} or ETH3D \cite{schops2017multi}, however, many of them turns out to be not adequate for this work purposes due to their high-quality images.
Indeed, applying Super-Resolution (SR) to high-resolution datasets would lead to at least 4K images, from which it would be unfeasible to reconstruct 3D models or even to apply the SR algorithm itself due to computational constraints.
Moreover, it would go beyond the scope of the paper, which explicitly addresses the low-resolution scenario.
For these reasons, we adopted the train and test datasets from the low-resolution many-view benchmark of ETH3D and the downsampled version of the train dataset from the Tanks and Temples benchmark.

\subsection{Evaluation over ETH3D Benchmark}
The low-resolution many-view ETH3D benchmark \cite{schops2017multi} is composed by 10 datasets of gray-scale images split by the authors in 5 sequences belonging to the train set $\mathcal{S}_{train}$ and 5 to the test set $\mathcal{S}_{test}$, each one divided in 3 Outdoor and 2 Indoor sequences. Each sequence $\mathcal{I}_{n}$ has its Pinhole camera model parameters $\mathcal{C}_{n}$ already estimated by the authors. To compute their equivalent for each Super-Resolution set $\mathcal{C}_{n}^{2}$, we multiply them by the scale factor $k=2$.
We also make a qualitative split between Textured and Textureless sequences to better investigate SR behaviour along an additional dimension. More in detail, we consider $delivery\_area$, $electro$, $terrains$ as Textureless train sequences and $forest$, $playground$ as Textured ones, while $storage\_room$, $storage\_room\_2$ as Textureless test sequences and $lakeside$, $sand\_box$, $tunnel$ as Textured ones.

For each adopted MVS algorithm, we reconstructed the default low-resolution sequences and their SR versions obtained via bicubic interpolation and DBPN.
We locally evaluated the train set with the PatchMatch-based MVS algorithms COLMAP and TAPA-MVS and with the deep-learning-based one CasMVSNet, comparing the F1, Accuracy and Completeness of the 3D models obtained.

\begin{figure*}[!tbp]
\begin{center}
\setlength{\tabcolsep}{1.25px}
\begin{tabular}{cccc}
\multicolumn{4}{c}{}\\
Model & Overall & Textured & Textureless\\

COLMAP & \raisebox{-.5\height}{\includegraphics[width=0.285\textwidth]{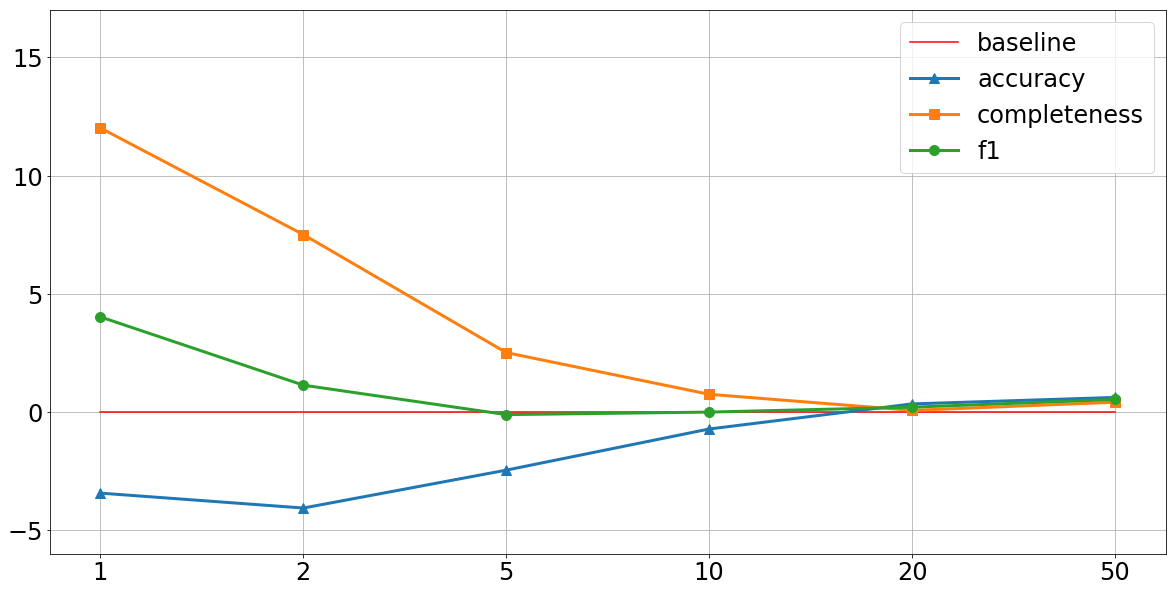}} & \raisebox{-.5\height}{\includegraphics[width=0.285\textwidth]{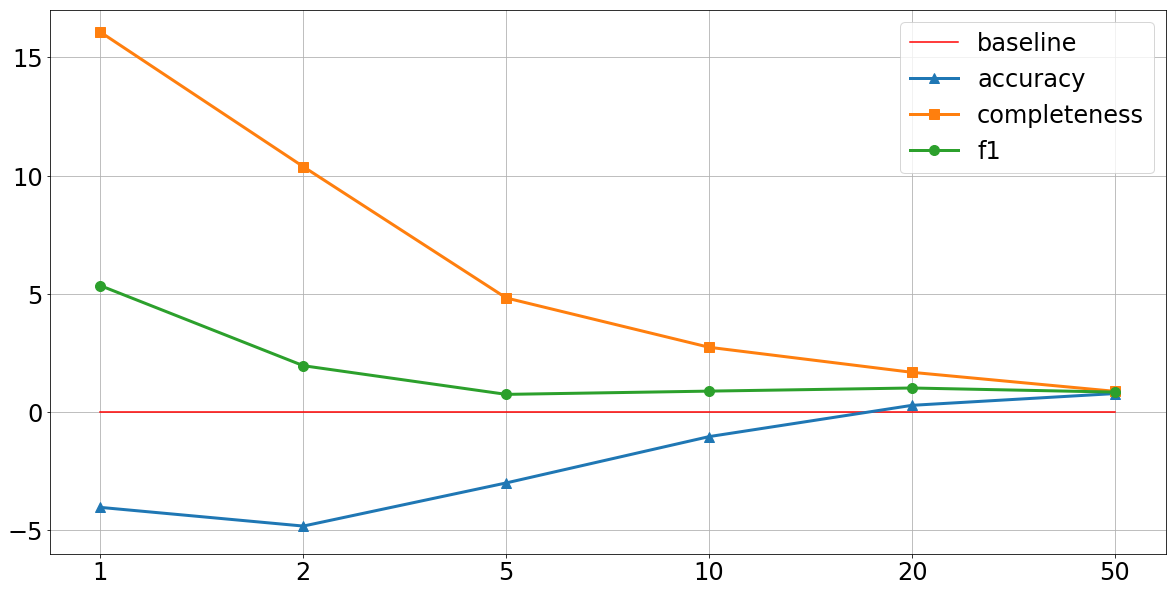}} &
\raisebox{-.5\height}{\includegraphics[width=0.285\textwidth]{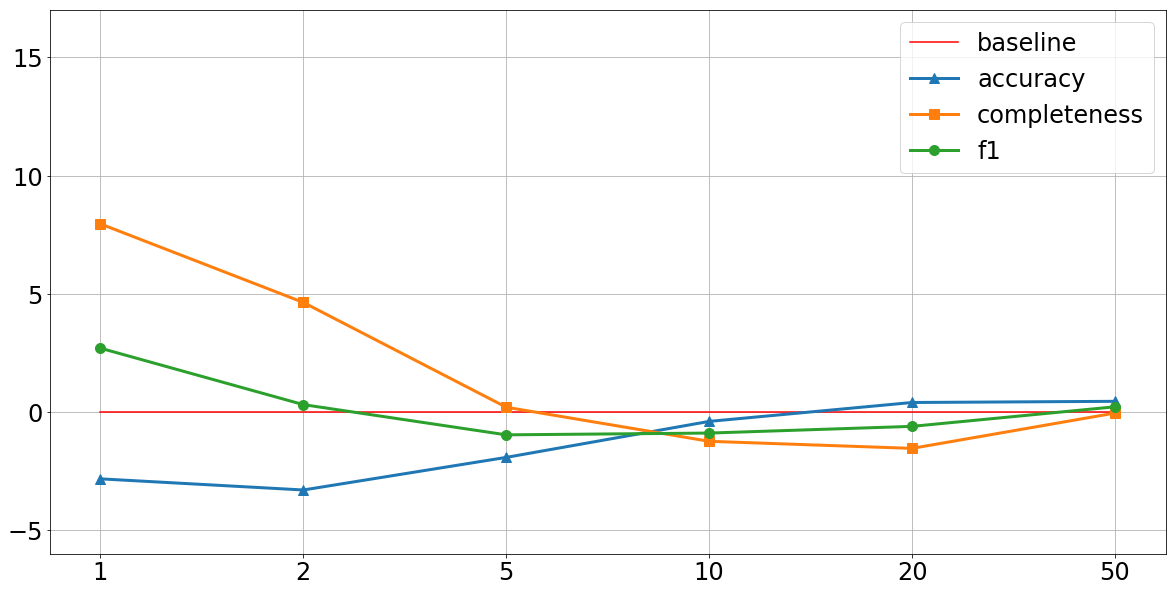}}\\

TAPA-MVS & \raisebox{-.5\height}{\includegraphics[width=0.285\textwidth]{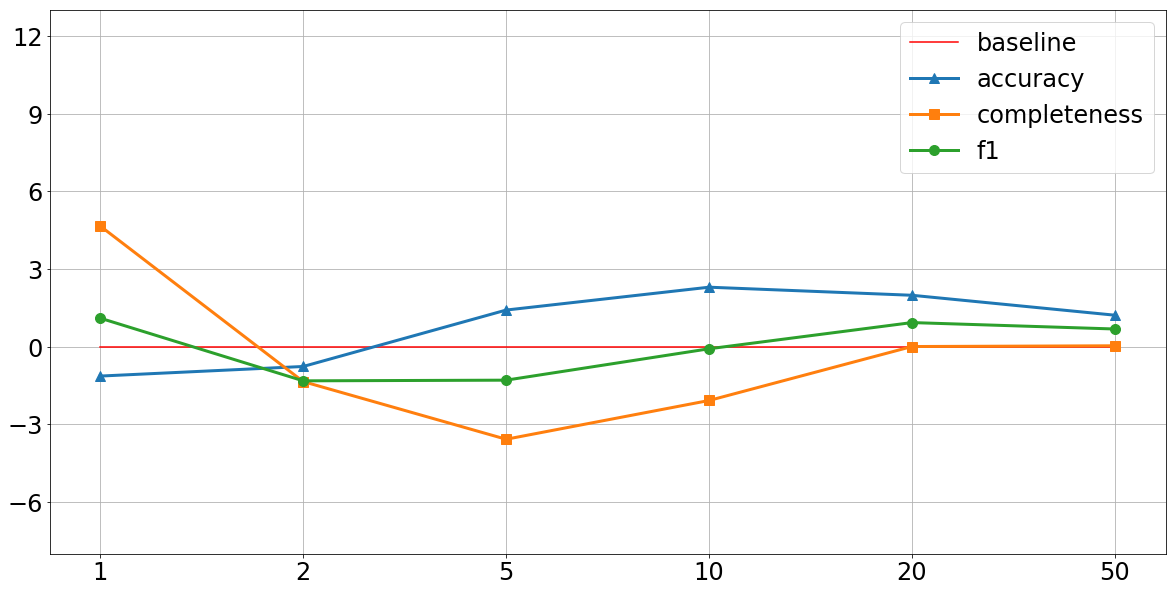}} & \raisebox{-.5\height}{\includegraphics[width=0.285\textwidth]{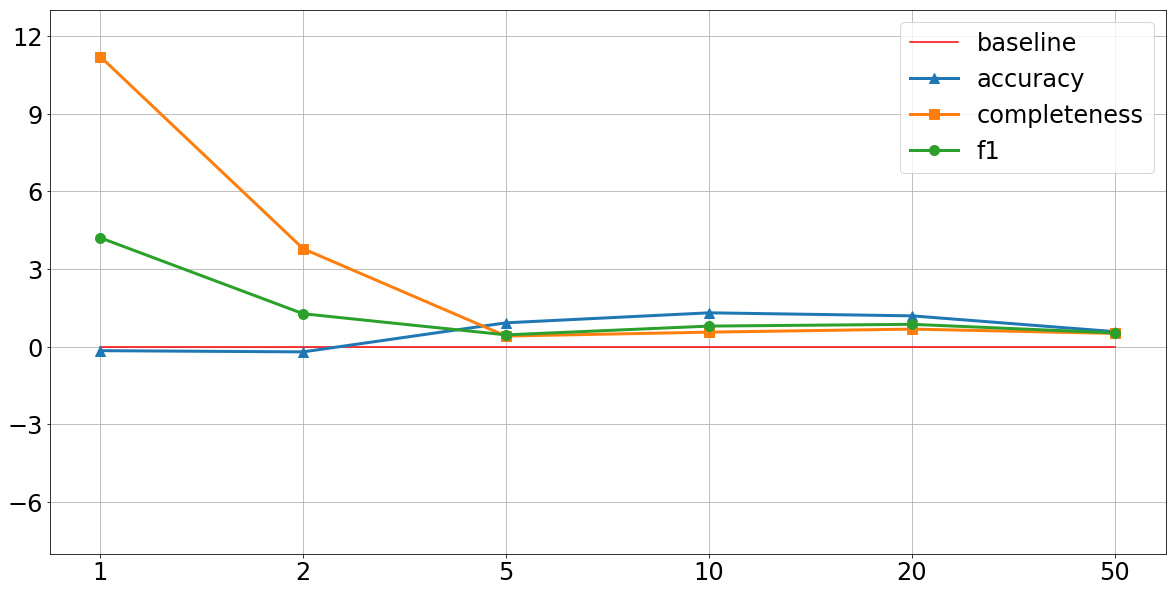}} &
\raisebox{-.5\height}{\includegraphics[width=0.285\textwidth]{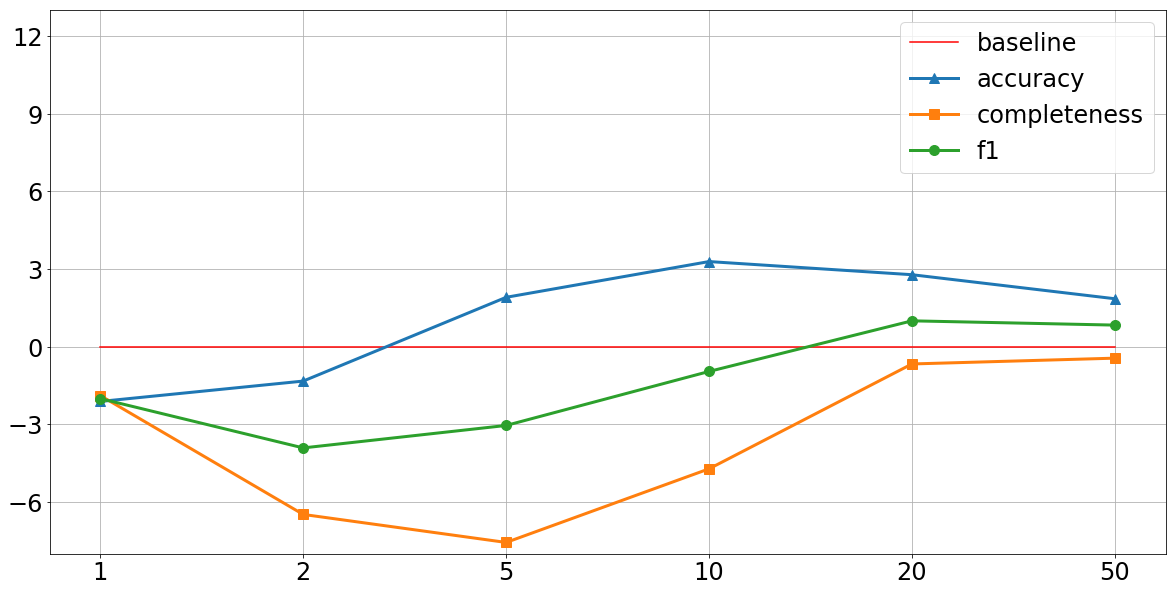}}\\

CasMVSNet & \raisebox{-.5\height}{\includegraphics[width=0.285\textwidth]{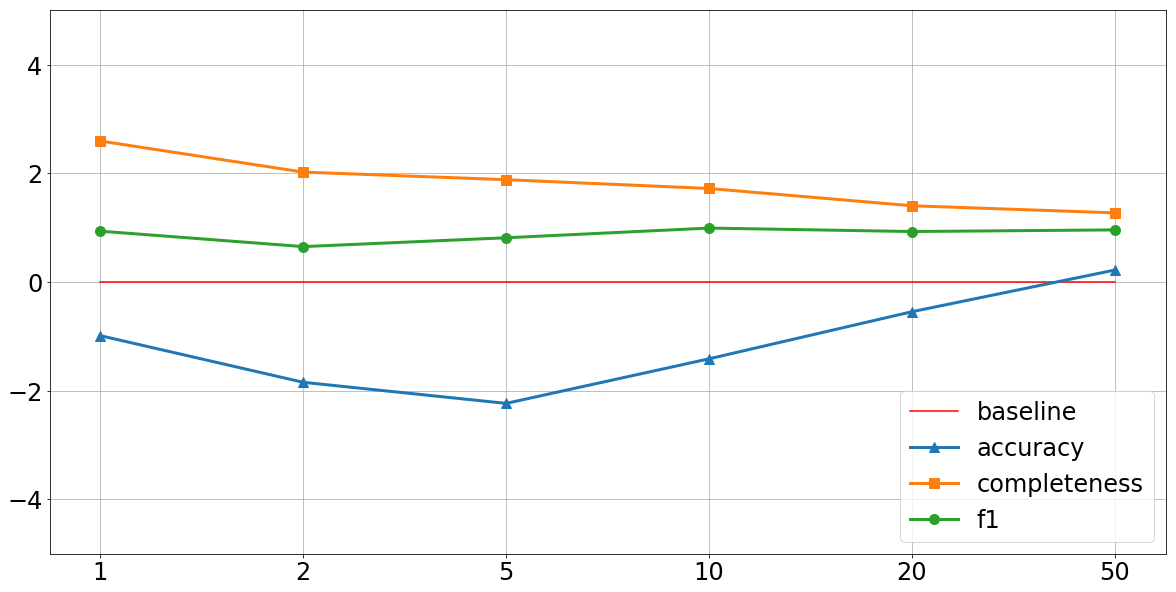}} & \raisebox{-.5\height}{\includegraphics[width=0.285\textwidth]{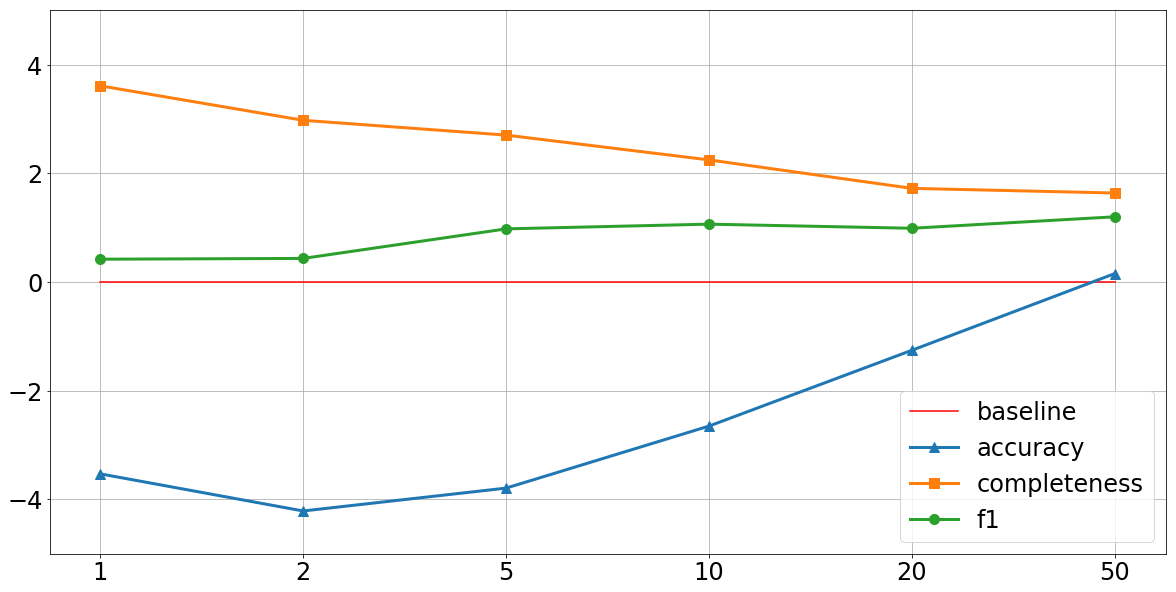}} &
\raisebox{-.5\height}{\includegraphics[width=0.285\textwidth]{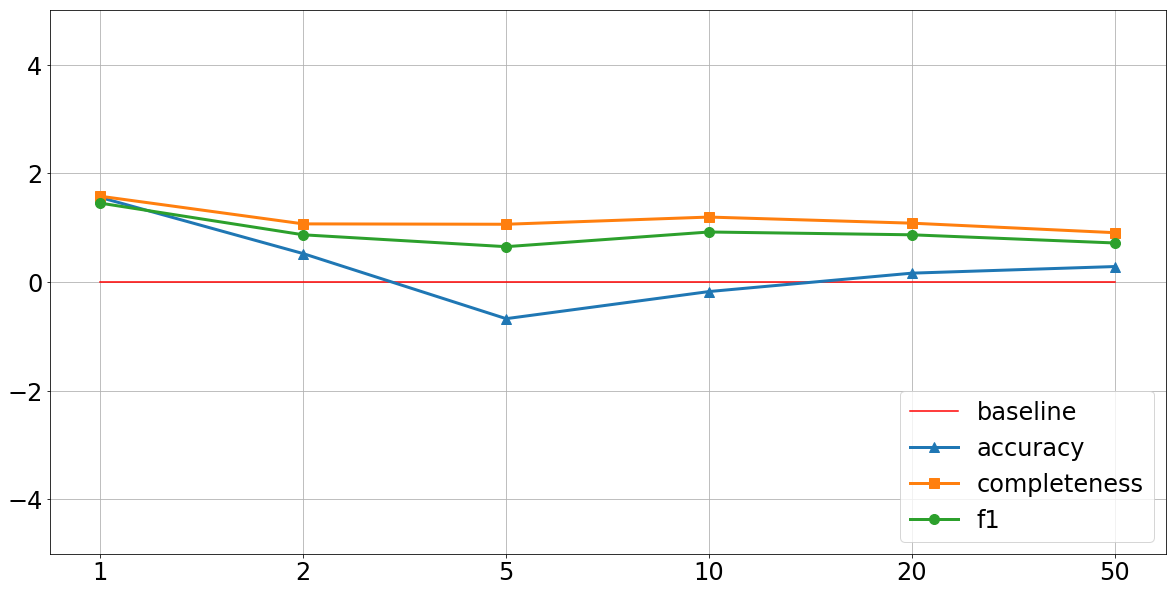}}\\

& \scriptsize{tolerance (cm)} & \scriptsize{tolerance (cm)} & \scriptsize{tolerance (cm)}\\
\end{tabular}
\caption{Improvements of F1, accuracy and completeness scores on the ETH3D low-resolution multi-view benchmark of proposed Super-Resolution models against their own low-resolution version, divided by the tolerances. On average, COLMAP Overall F1 improvement is 0.96\%, the Textureless is 0.13\%, the Textured is 1.80\%, TAPA-MVS Overall F1 improvement is 0.01\%, the Textureless is -1.35\%, the Textured is 1.36\% and CasMVSNet Overall F1 improvement is 0.88\%, the Textureless is 0.91\%, the Textured is 0.85\%.}
\label{fig:final_plot}
\end{center}
\end{figure*}

Table \ref{tab:COLMAPtrain} shows that in Overall both SISR algorithms allow COLMAP to reconstruct better 3D models, especially while considering more stringent evaluation criteria, i.e. small tolerances.
The most notable SR improvement is appreciable in textured and outdoor sequences in which, as shown in the section (a) of Figure \ref{fig:depthmapscolmap}, the algorithm is able to compute deeper depth maps thanks to the increased amount of input information.
In textureless sequences, which in this case are almost overlapped with indoor ones, COLMAP, as expected, is not able to gain the same improvements as before.
Nevertheless, despite the algorithm nature and SR artifacts, as shown in the section (b) of Figure \ref{fig:depthmapscolmap} it is able to produce better depth estimates near the perimeters and in the more detailed areas while producing more holes in the areas with less information.

Table \ref{tab:TAPAtrain} shows how TAPA-MVS, being a COLMAP extension to better handle textureless regions, follows the same COLMAP behaviour to SR input in textured and outdoor sequences, while produces poorer results in the others.
This behaviour is amenable to the presence of SISR artifacts in the input images which do not allow TAPA-MVS to propagate its depth hypothesis to the textureless area.
Despite this drawback, SISR lets the algorithm improve its overall performance especially for small tolerances.

Table \ref{tab:CasMVSNettrain} shows how CasMVSNet, the deep-learning-based MVS algorithm, reacts differently from the previous: it is able to exploit the increased information in input improving both textured and textureless performance, especially for small tolerances, although in a more attenuated way compared to the previously analyzed PatchMatch-based algorithms.
The SR application turns out to be overall effective with both the SISR algorithms.

\begin{table*}[tbp]
    \caption{F1, accuracy and completeness scores over ETH3D low-resolution multi-view benchmark. We compare the presented models grouped in many subsets with a tolerance $\tau = 1 cm$}
    \label{tab:finaltable}
    \centering
    \setlength{\tabcolsep}{1px}
    \begin{tabular}{c|ccc|ccc|ccc|ccc|ccc|ccc|ccc}
    {model} & \multicolumn{3}{c}{Overall} & \multicolumn{3}{c}{Train} & \multicolumn{3}{c}{Test} & \multicolumn{3}{c}{Indoor} & \multicolumn{3}{c}{Outdoor}& \multicolumn{3}{c}{Textured} & \multicolumn{3}{c}{Textureless}\\
    
    & F1 & acc & comp & F1 & acc & comp & F1 & acc & comp & F1 & acc & comp & F1 & acc & comp & F1 & acc & comp & F1 & acc & comp \\

    COLMAP   & 36.6 & $\mathbf{40.7}$ & 33.8    & 35.8 & $\mathbf{39.4}$ & 33.4   & 37.4 & $\mathbf{42.0}$ & 34.1   & 34.4 & $\mathbf{38.1}$ & 31.7   & 38.0 & $\mathbf{42.4}$ & 35.1   & 38.9 & $\mathbf{42.9}$ & 36.3   & 34.2 & $\mathbf{38.4}$ & 31.2\\
    COLMAP (DBPN)   & $\mathbf{40.6}$ & 37.2 & $\mathbf{45.8}$    & $\mathbf{40.0}$ & 36.97 & $\mathbf{45.2}$   & $\mathbf{41.2}$ & 37.5 & $\mathbf{46.3}$   & $\mathbf{36.5}$ & 35.7 & $\mathbf{38.2}$   & $\mathbf{43.3}$ & 38.2 & $\mathbf{50.8}$   & $\mathbf{44.3}$ & 38.9 & $\mathbf{52.4}$   & $\mathbf{36.9}$ & 35.6 & $\mathbf{39.1}$\\
    \cline{1-22}
    TAPA   & 40.4 & $\mathbf{41.8}$ & 40.3    & 38.9 & $\mathbf{42.0}$ & 38.2   & $\mathbf{42.0}$ & $\mathbf{41.5}$ & 42.4   & $\mathbf{41.4}$ & 41.6 & $\mathbf{41.6}$   & 39.8 & $\mathbf{41.9}$ & 39.5   & 40.9 & $\mathbf{40.2}$ & 42.2   & $\mathbf{40.0}$ & $\mathbf{43.4}$ & $\mathbf{38.5}$\\
    TAPA-MVS (DBPN)   & $\mathbf{41.5}$ & 40.7 & $\mathbf{45.0}$    & $\mathbf{42.3}$ & 41.1 & $\mathbf{45.3}$   & 40.7 & 40.2 & $\mathbf{44.6}$   & 37.0 & $\mathbf{41.9}$ & 34.3   & $\mathbf{44.5}$ & 39.8 & $\mathbf{52.1}$   & $\mathbf{45.1}$ & 40.0 & $\mathbf{53.4}$   & 38.0 & 41.3 & 36.6\\
    \cline{1-22}
    CasMVSNet   & 36.8 & $\mathbf{42.4}$ & 34.3    & 38.3 & $\mathbf{44.7}$ & 34.9   & 35.3 & $\mathbf{40.0}$ & 33.6   & 27.8 & 36.0 & 23.7   & 42.7 & $\mathbf{46.8}$ & 41.3   & 43.2 & $\mathbf{47.2}$ & 42.1   & 30.4 & 37.5 & 26.4\\
    CasMVSNet (DBPN)   & $\mathbf{37.7}$ & 41.4 & $\mathbf{36.9}$    & $\mathbf{39.6}$ & 44.6 & $\mathbf{37.1}$   & $\mathbf{35.9}$ & 38.2 & $\mathbf{36.7}$   & $\mathbf{28.9}$ & $\mathbf{37.4}$ & $\mathbf{24.7}$   & $\mathbf{43.6}$ & 44.0 & $\mathbf{45.0}$   & $\mathbf{43.6}$ & 43.6 & $\mathbf{45.7}$   & $\mathbf{31.8}$ & $\mathbf{39.1}$ & $\mathbf{28.0}$\\
    \end{tabular}
\end{table*}

So far DBPN led on average to better results than bicubic SR, thus we have chosen the former to evaluate SISR effects over the test set which needed to be publicly evaluated over ETH3D leaderboard and we could not access directly.

From Figure \ref{fig:final_plot} and Table \ref{tab:finaltable} we can assert that, although with some bias, DBPN SISR makes MVS algorithms perform according with the scores over the train set: COLMAP gains remarkable improvements for low tolerances, improving especially in reconstructing textured sequences.
TAPA-MVS improves regardless the considered tolerance in textured sequences, while is not able to improve for textureless sequences due to its specific nature which makes it not compatible with SISR in this specific scenario.
Finally, CasMVSNet is always able to improve thanks to DBPN despite the tolerance considered.

\begin{figure*}[p]
\begin{center}
\setlength{\tabcolsep}{0.3125px}
\begin{tabular}{cccccc}
\multicolumn{6}{c}{electro}\\
\includegraphics[width=0.16\textwidth]{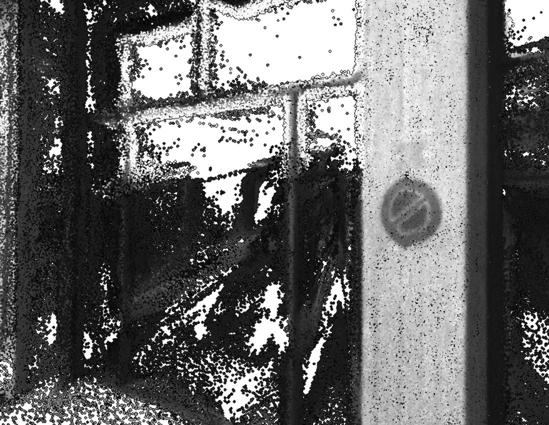}&
\includegraphics[width=0.16\textwidth]{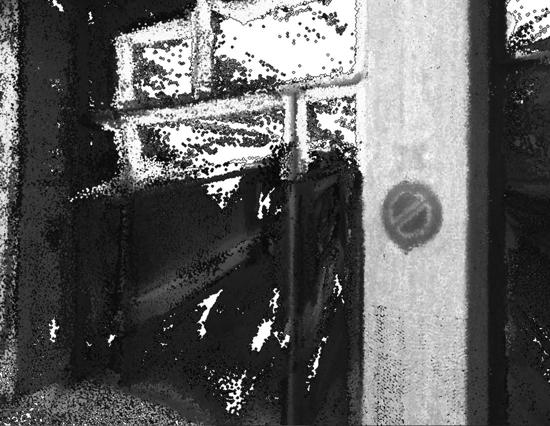}&
\includegraphics[width=0.16\textwidth]{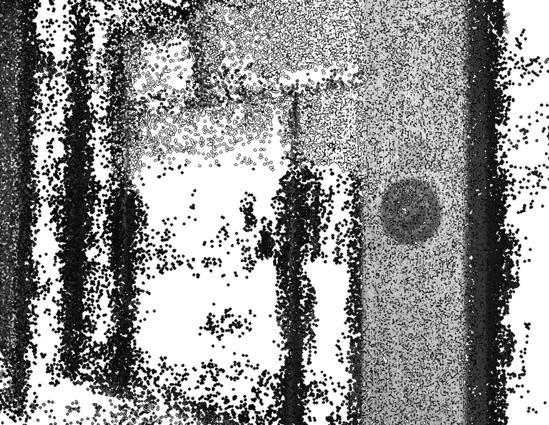}&
\includegraphics[width=0.16\textwidth]{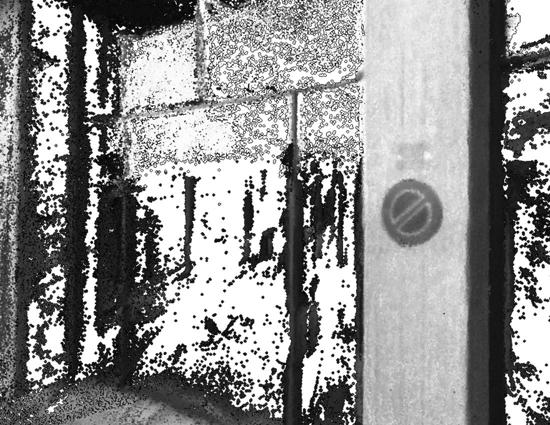}&
\includegraphics[width=0.16\textwidth]{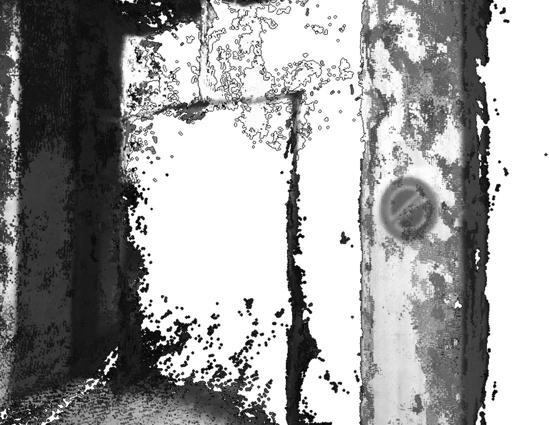}&
\includegraphics[width=0.16\textwidth]{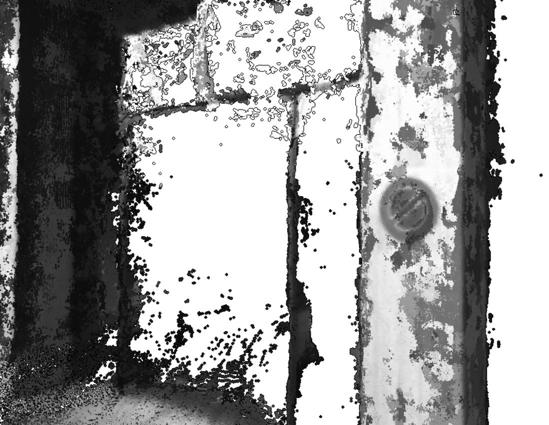}\\
\multicolumn{6}{c}{forest}\\
\includegraphics[width=0.16\textwidth]{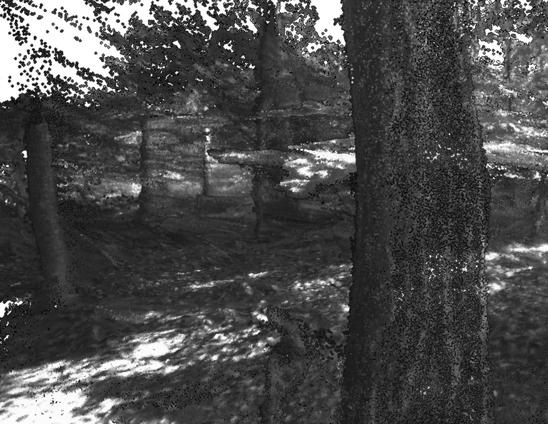}&
\includegraphics[width=0.16\textwidth]{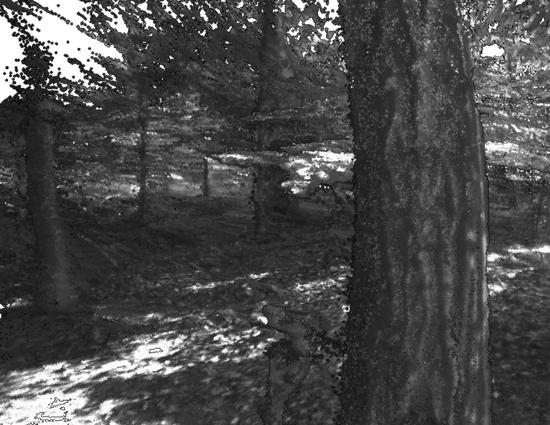}&
\includegraphics[width=0.16\textwidth]{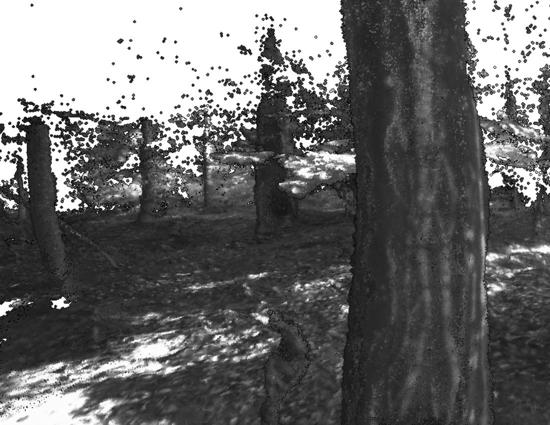}&
\includegraphics[width=0.16\textwidth]{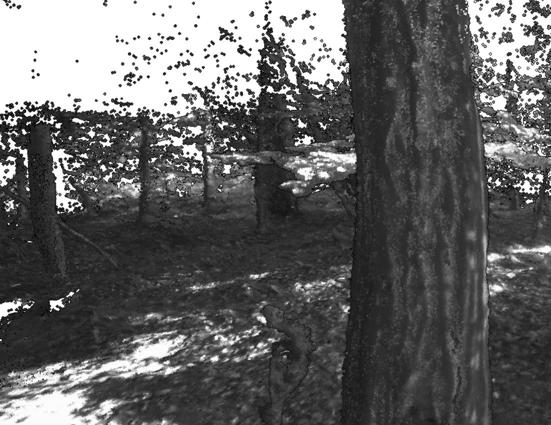}&
\includegraphics[width=0.16\textwidth]{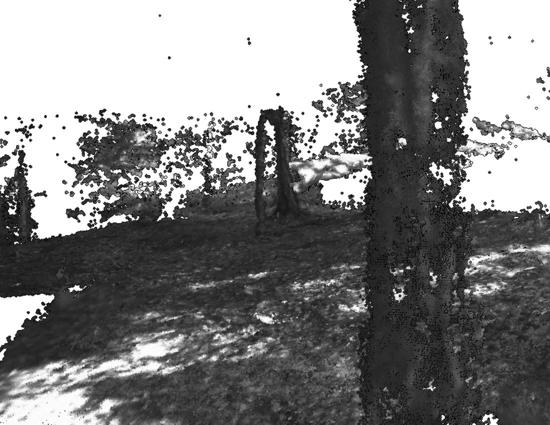}&
\includegraphics[width=0.16\textwidth]{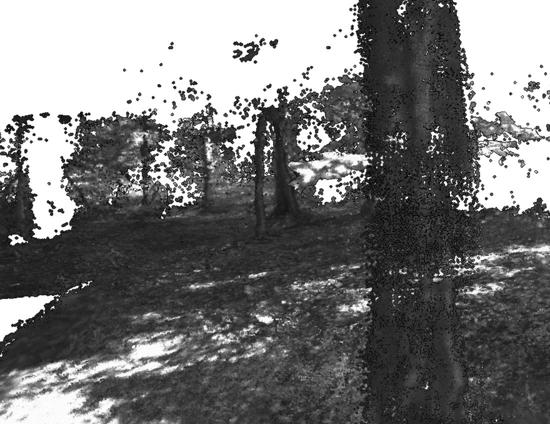}\\
\multicolumn{6}{c}{playground}\\
\includegraphics[width=0.16\textwidth]{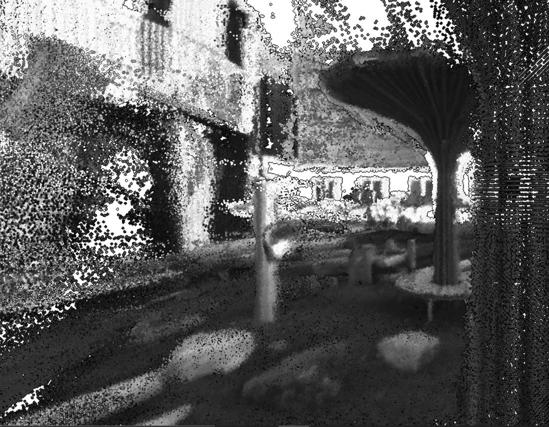}&
\includegraphics[width=0.16\textwidth]{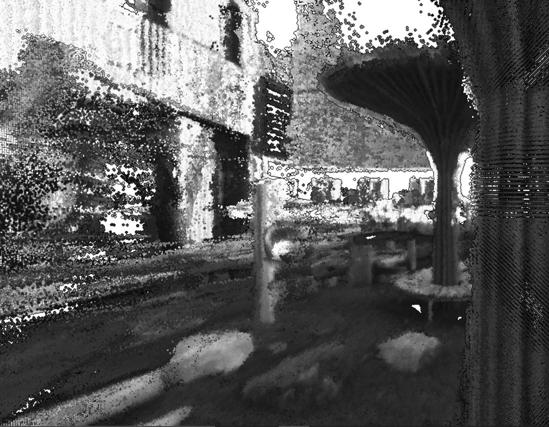}&
\includegraphics[width=0.16\textwidth]{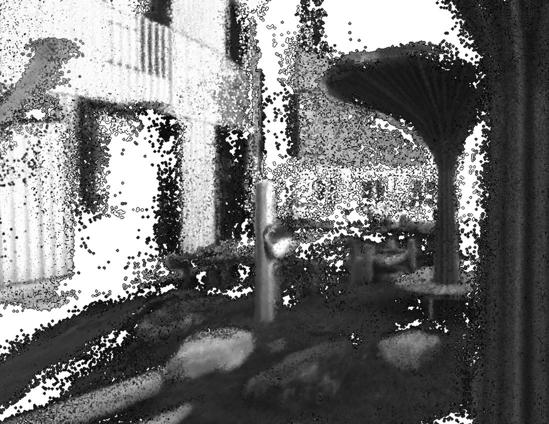}&
\includegraphics[width=0.16\textwidth]{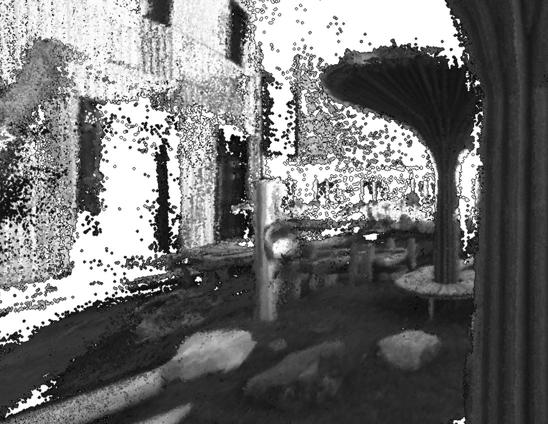}&
\includegraphics[width=0.16\textwidth]{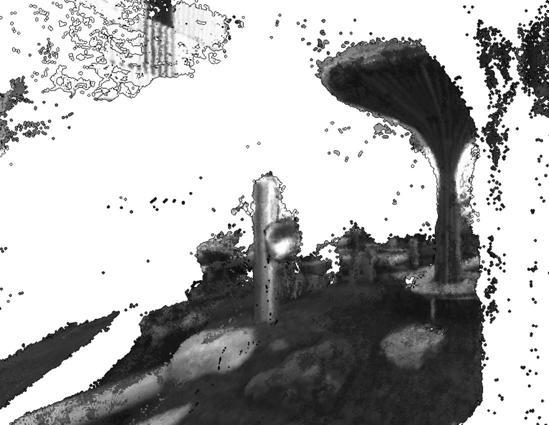}&
\includegraphics[width=0.16\textwidth]{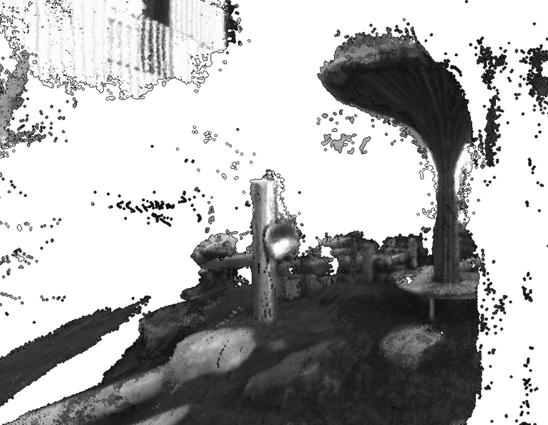}\\
\multicolumn{6}{c}{lakeside}\\
\includegraphics[width=0.16\textwidth]{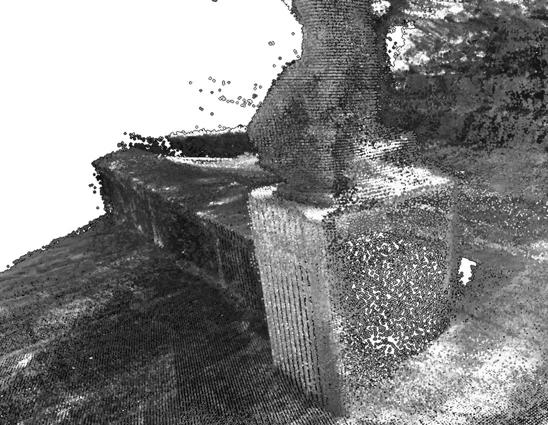}&
\includegraphics[width=0.16\textwidth]{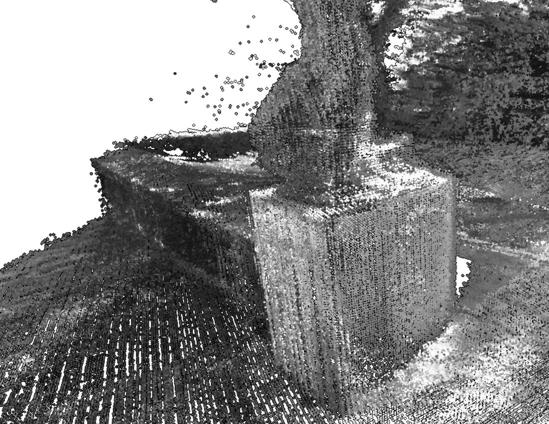}&
\includegraphics[width=0.16\textwidth]{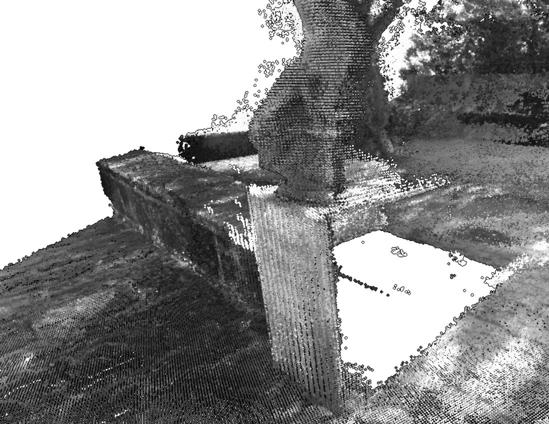}&
\includegraphics[width=0.16\textwidth]{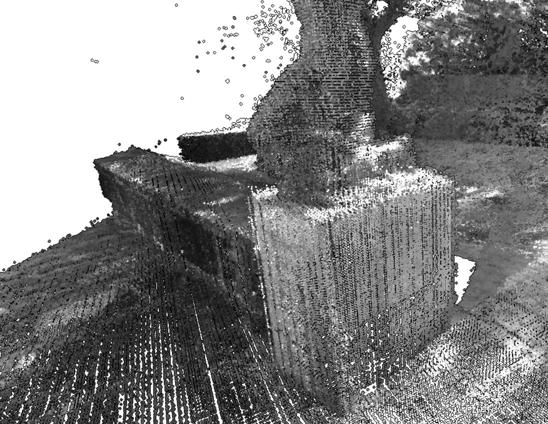}&
\includegraphics[width=0.16\textwidth]{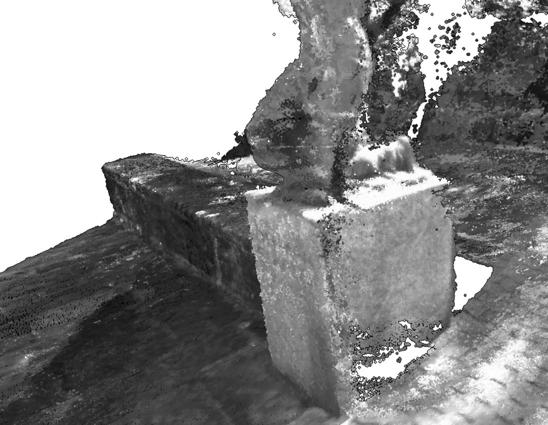}&
\includegraphics[width=0.16\textwidth]{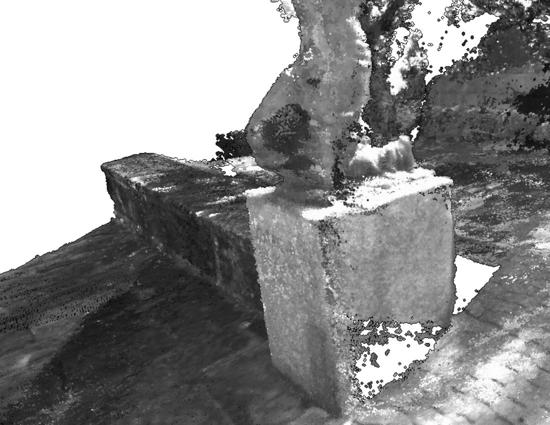}\\
\multicolumn{6}{c}{sand\_box}\\
\includegraphics[width=0.16\textwidth]{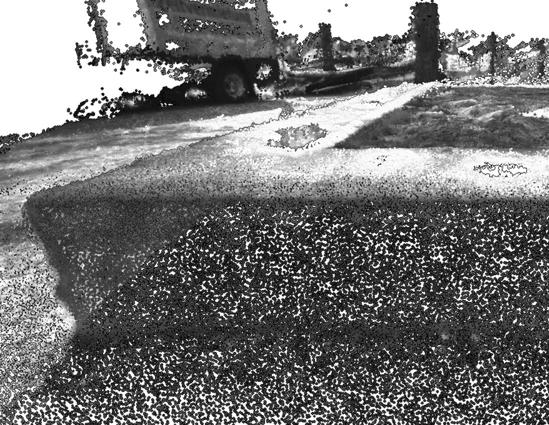}&
\includegraphics[width=0.16\textwidth]{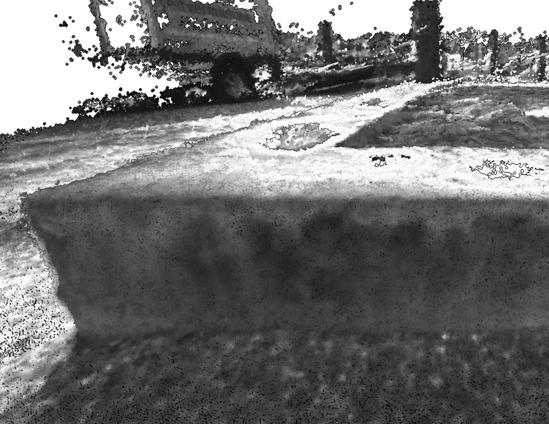}&
\includegraphics[width=0.16\textwidth]{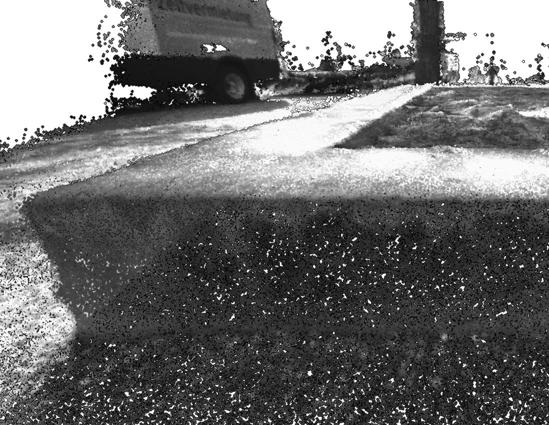}&
\includegraphics[width=0.16\textwidth]{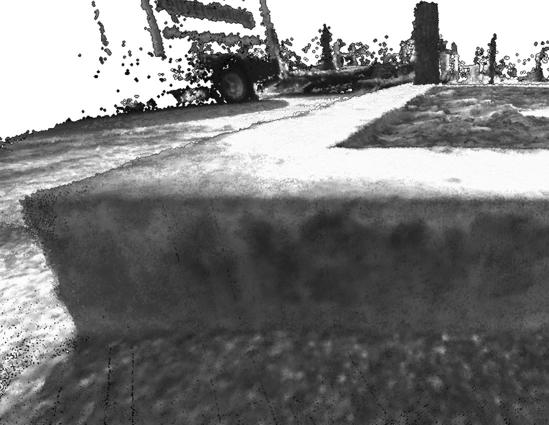}&
\includegraphics[width=0.16\textwidth]{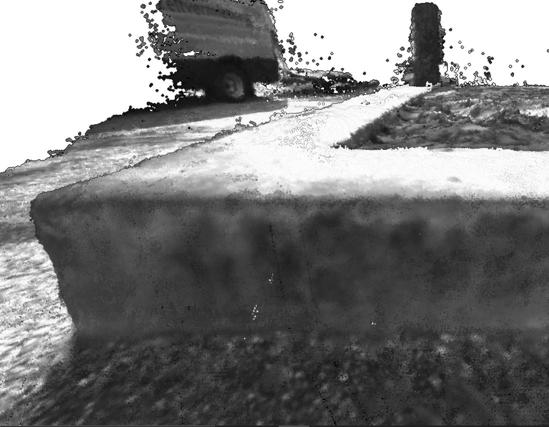}&
\includegraphics[width=0.16\textwidth]{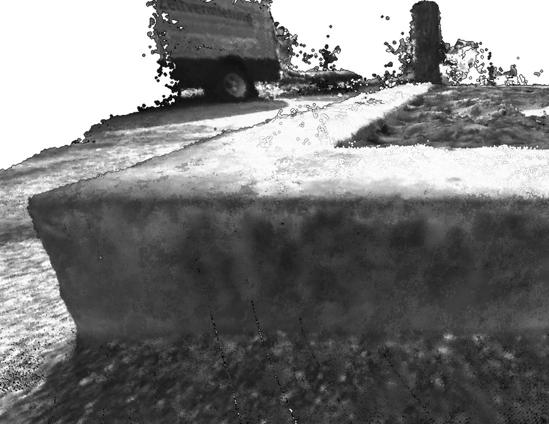}\\
\multicolumn{6}{c}{storage\_room}\\
\includegraphics[width=0.16\textwidth]{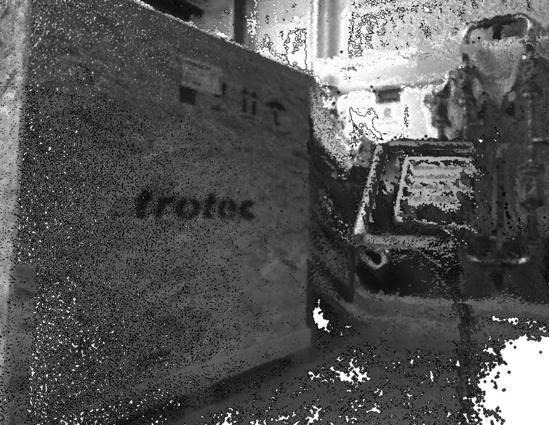}&
\includegraphics[width=0.16\textwidth]{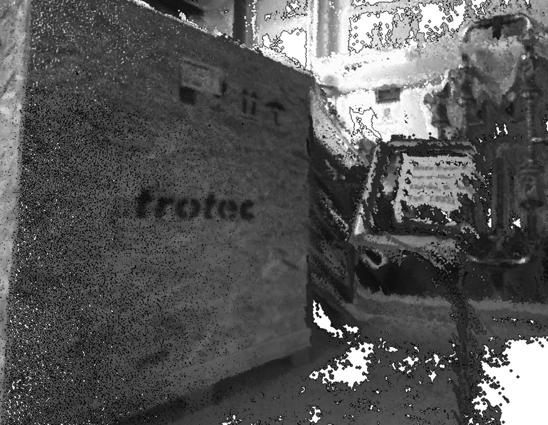}&
\includegraphics[width=0.16\textwidth]{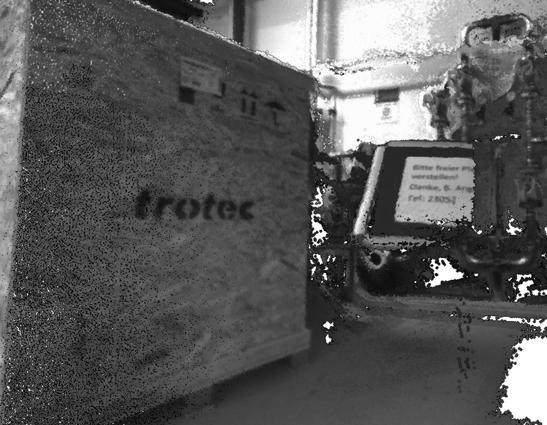}&
\includegraphics[width=0.16\textwidth]{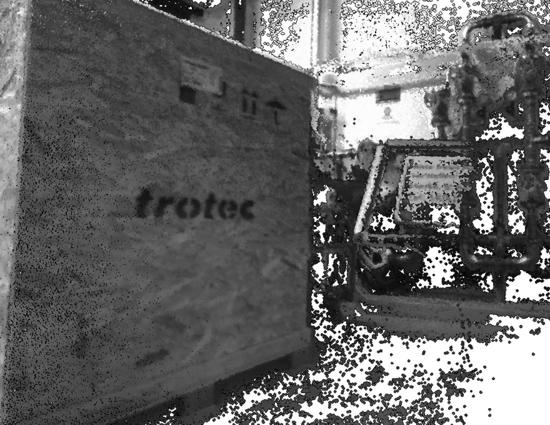}&
\includegraphics[width=0.16\textwidth]{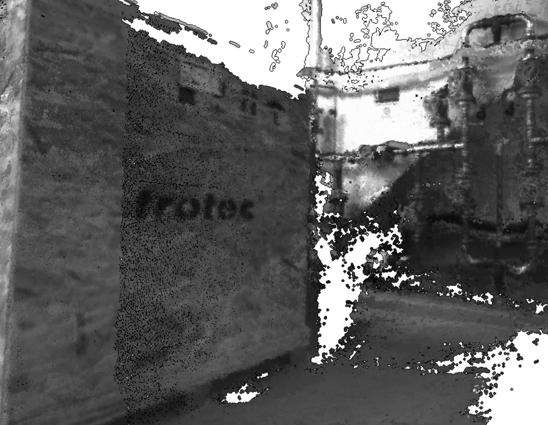}&
\includegraphics[width=0.16\textwidth]{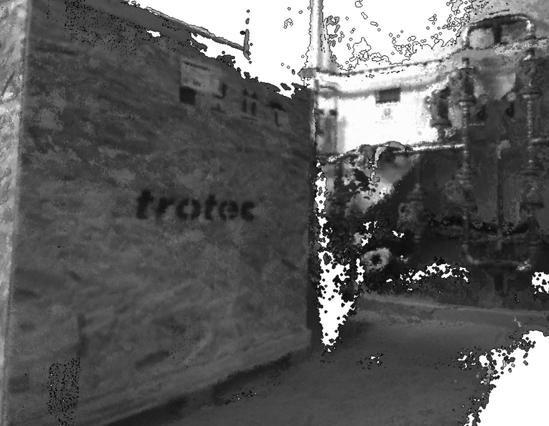}\\
\multicolumn{6}{c}{storage\_room\_2}\\
\includegraphics[width=0.16\textwidth]{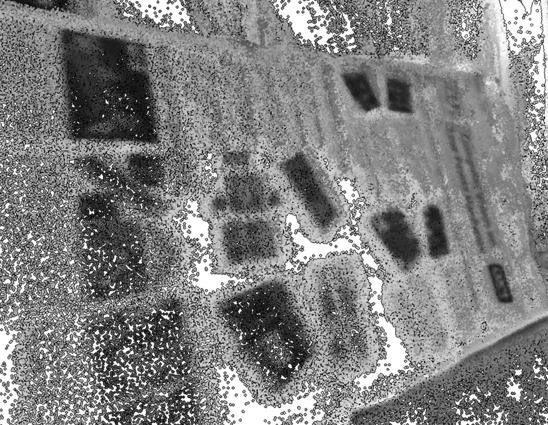}&
\includegraphics[width=0.16\textwidth]{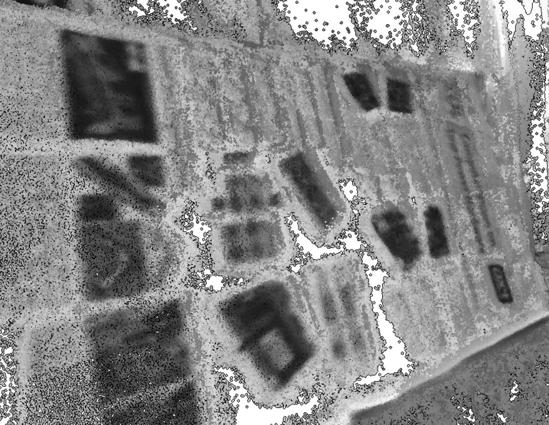}&
\includegraphics[width=0.16\textwidth]{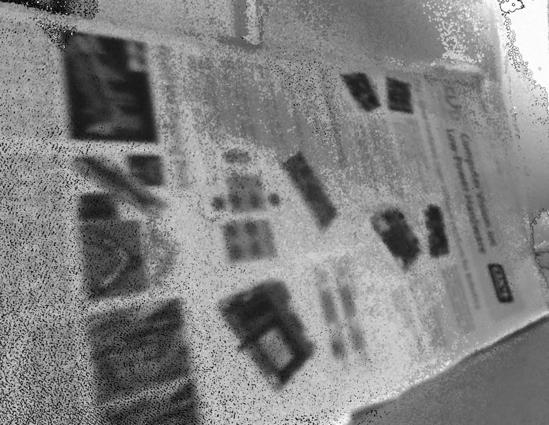}&
\includegraphics[width=0.16\textwidth]{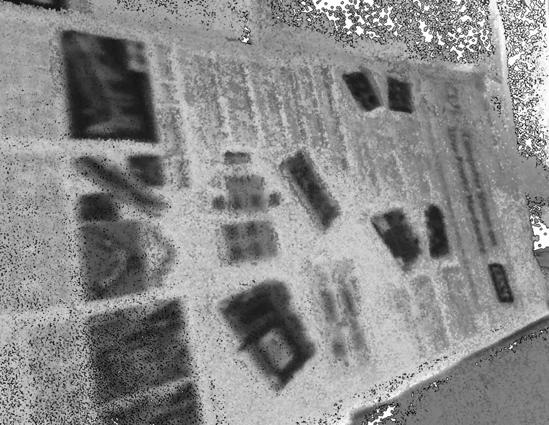}&
\includegraphics[width=0.16\textwidth]{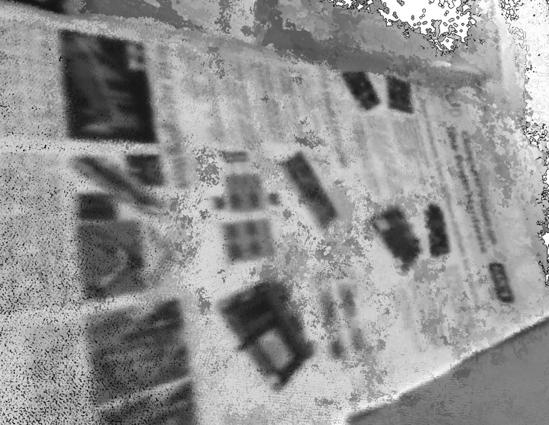}&
\includegraphics[width=0.16\textwidth]{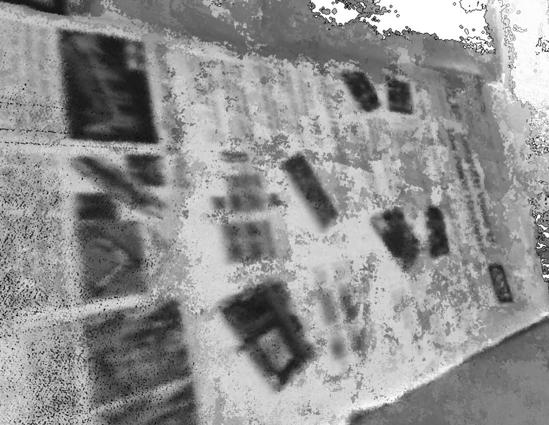}\\
\multicolumn{6}{c}{tunnel}\\
\includegraphics[width=0.16\textwidth]{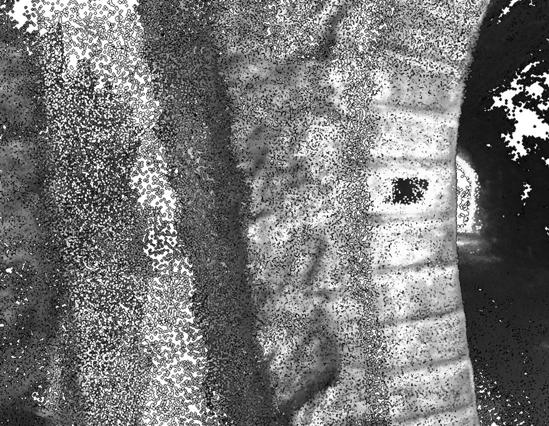}&
\includegraphics[width=0.16\textwidth]{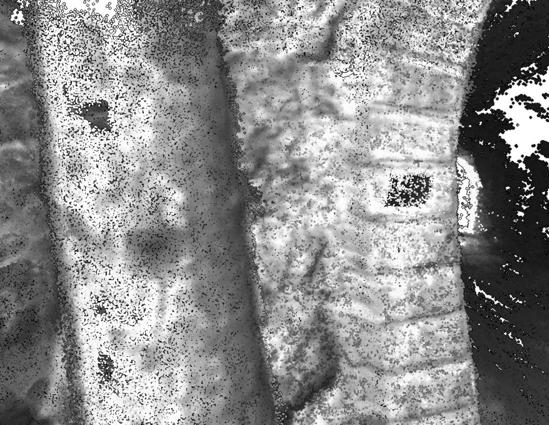}&
\includegraphics[width=0.16\textwidth]{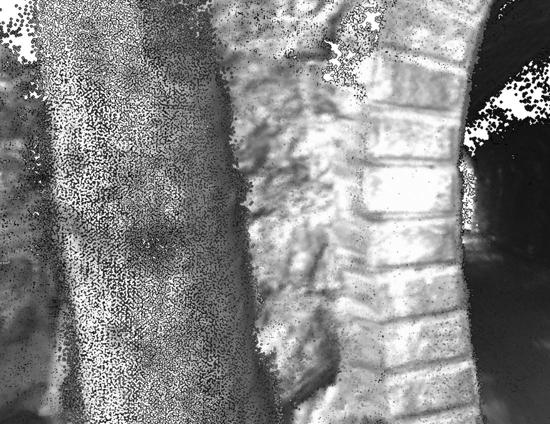}&
\includegraphics[width=0.16\textwidth]{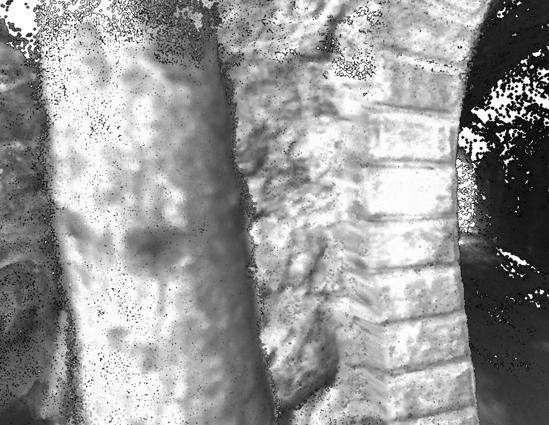}&
\includegraphics[width=0.16\textwidth]{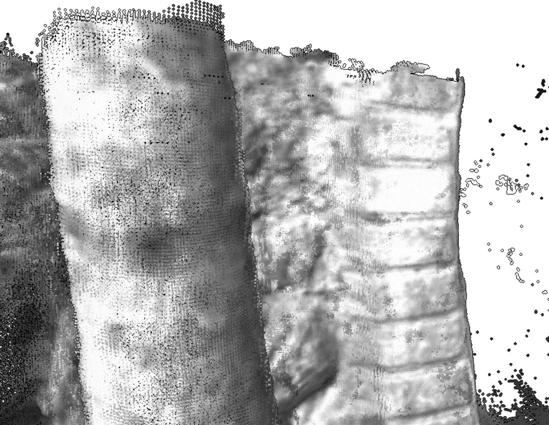}&
\includegraphics[width=0.16\textwidth]{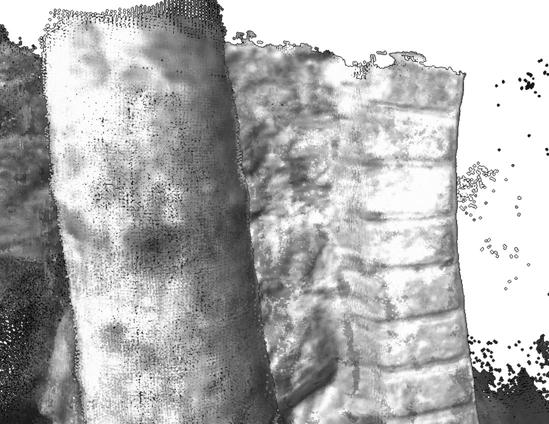}\\

COLMAP & COLMAP (DBPN) & TAPA-MVS & TAPA-MVS (DBPN) & CasMVSNet & CasMVSNet (DBPN)\\
\end{tabular}
    \vspace*{5pt}
\caption{Details of ETH3D low-res many-view benchmark 3D reconstructions. We compare the same view for each proposed pipeline in both low- and DBPN-SR- versions.}
\label{fig:eth3D_res}
\end{center}
\end{figure*}

According to the adopted metrics, it is evident how SISR is able to affect positively MVS pipelines with a positive trade-off between completeness and accuracy, as can be seen from Figure \ref{fig:eth3D_res} in which the 3D models obtained from SR sequences results much more dense and rich in details.

More in detail, as shown in Table \ref{tab:finaltable}, SISR effect upon MVS pipelines has its major impact over completeness when considering the more stringent evaluation criteria, i.e., $\tau=1cm$.
This means not only that the increased number of pixel in input images is effectively mapped to an increased number of points in the 3D space, but also that the latter are particularly thickened near the points of the corresponding ground truth reconstructions.
From these results we can assert SISR is strongly recommended while reconstructing Textured sequences starting from low-resolution images, being able to highlight details, which, in turn, leads to better depth estimations and to an increased number of matches during stereo fusion.

The 3D models evaluated and the scores showed so far are publicly available for further inspections at \url{https://www.eth3d.net/low\_res\_many\_view} with the names COLMAP(base), COLMAP(SR), TAPA-MVS, TAPA-MVS(SR), CasMVSNet(base), CasMVSNet(SR\_A) and CasMVSNet(SR\_B) due to limits in maximum upload space.

\begin{figure*}[tbp]
    \centering
    \setlength{\tabcolsep}{1.25px}
    \begin{tabular}{ccc}
    \multicolumn{3}{c}{}\\
    \footnotesize{Overall} & \footnotesize{Outdoor} & \footnotesize{Indoor}\\
    \includegraphics[width=0.325\textwidth]{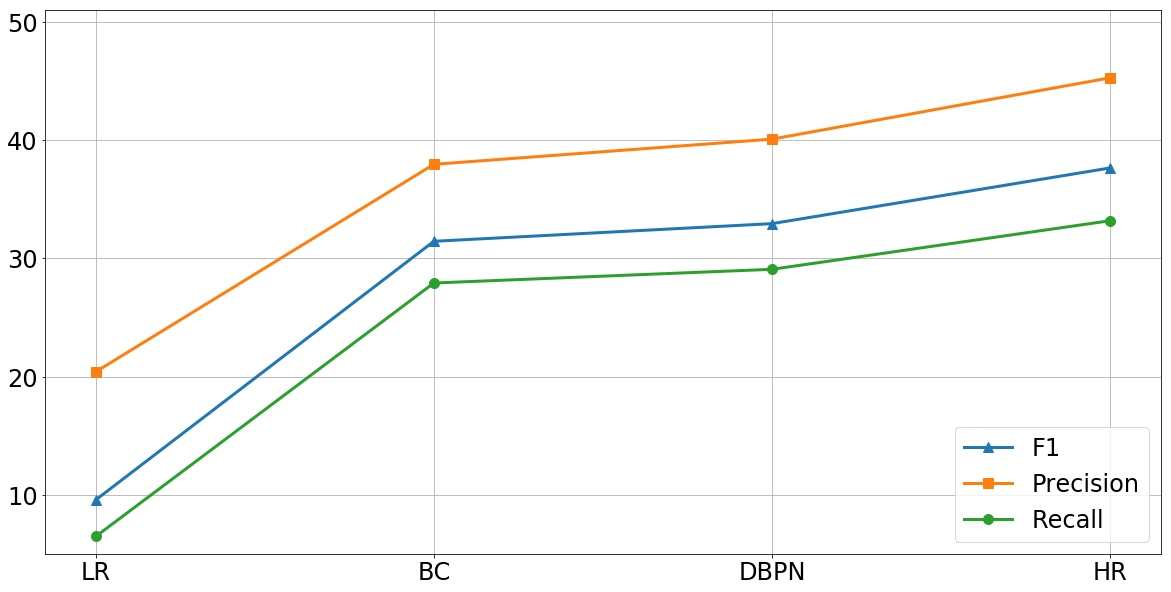}&
    \includegraphics[width=0.325\textwidth]{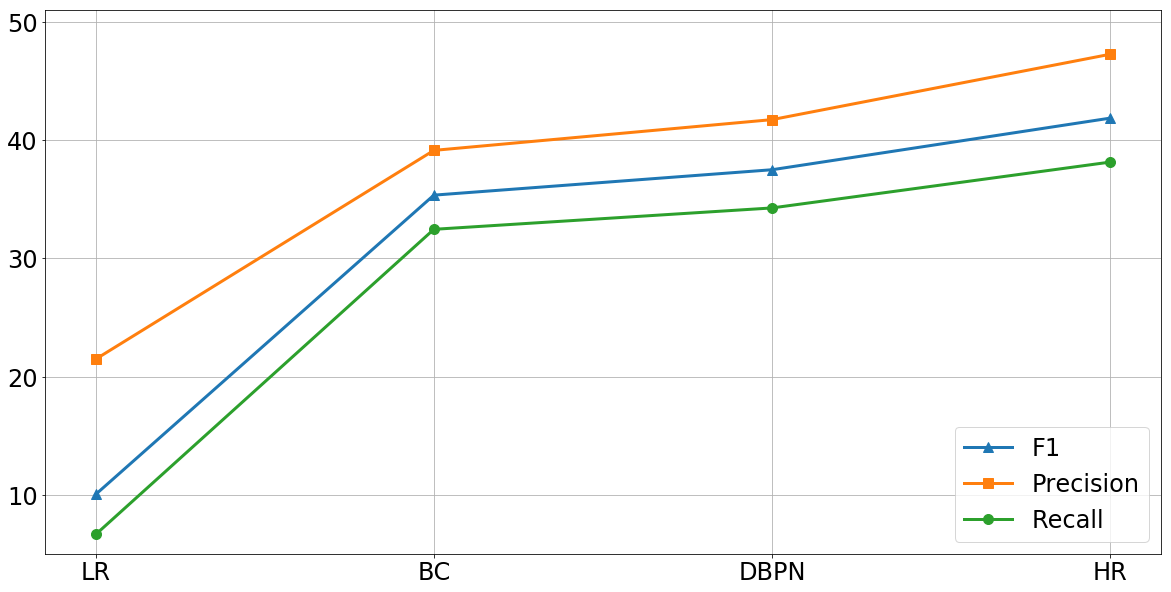}&
    \includegraphics[width=0.325\textwidth]{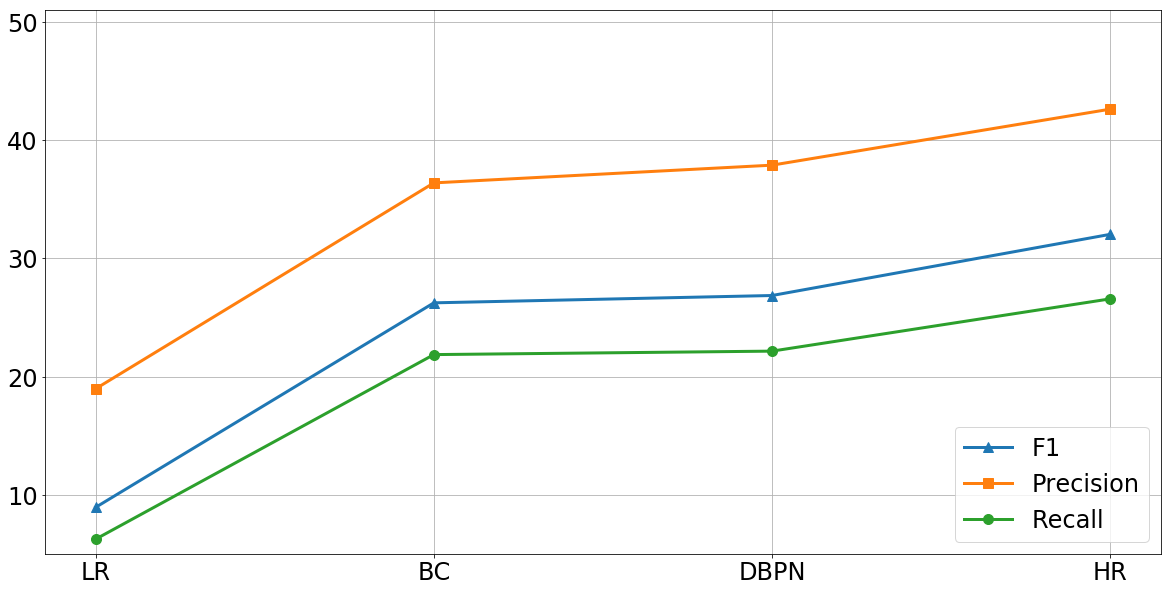}\\
    \end{tabular}
    \caption{COLMAP average scores over Tanks and Temples train benchmark down-sampled sequences grouped by Overall, Outdoor and Indoor sequences. In detail, 3D reconstruction F1, Precision and Recall scores starting from $\mathcal{S}_{LR}$, $\mathcal{S}_{BC}$, $\mathcal{S}_{DBPN}$ and $\mathcal{S}_{HR}$ are shown.
    On average, the F1 score of DBPN sequences improves w.r.t. LR ones by 23.38\%, 27.47\% and 17.92\% in Overall, Outdoor and Indoor sequences respectively.
    On average, the F1 score of BC sequences improves w.r.t. LR ones by 21.89\%, 25.32\% and 17.30\% in Overall, Outdoor and Indoor sequences respectively.}
    \label{fig:tet}
\end{figure*}

\subsection{Evaluation over Tanks and Temples Benchmark}
The Tanks and Temples train benchmark \cite{knapitsch2017tanks} is composed by 7 RGB textured datasets of high-resolution images. 
Due to computational constraints, we could not apply SR directly on them, thus we applied a bicubic down-sampling with scale factor 1/4 so that $\mathcal{S}_{LR} = {BC}_{1/4}(\mathcal{S})$ is the low-resolution set. In the same way, we computed the high-resolution set $\mathcal{S}_{HR} = {BC}_{1/2}(\mathcal{S})$ which can be considered as the ground truth for our SISR task.
Finally, we created the super-resolution sets $\mathcal{S}_{BC} = {BC}_{2}(\mathcal{S}_{LR})$ and $\mathcal{S}_{DL} = {DBPN}_{2}(\mathcal{S}_{LR})$.
For each sequence of each set we used COLMAP to compute camera parameters and reconstruct the 3D models. 
Then we evaluated F1, Precision and Recall metrics w.r.t. laser scans ground truth and a different tolerance for each sequence according with the benchmark authors.

From Figure \ref{fig:tet}, it is evident how the reconstructions starting from $\mathcal{S}_{HR}$ achieve on average a remarkable performance boost compared with $\mathcal{S}_{LR}$ ones. This result implies that in a scenario with a low-quality input the 3D output is strongly related with the images resolution.
Furthermore, regardless the reconstruction object being Outdoor or Indoor, SR is always able to improve the scores, obtaining scores much higher tank low-resolution ones.
At the same time, both the reconstructions starting from $\mathcal{S}_{BC}$ and $\mathcal{S}_{DL}$ achieve performance much close to the high-resolution reconstructions considered as ground truth for this evaluation, demonstrating that in this scenario both bicubic interpolation and DBPN artifacts do not have an incisive impact on the final outcome.
Despite the remarkable Precision improvement, SISR turns out to have positively impacted Recall majorly; this means SR 3D models have been reconstructed with a higher amount of points correctly estimated near the ground truth. 
Also in this case, DBPN SISR is able to produce on average better images for the MVS task w.r.t. bicubic interpolation without fine-tuning and being trained on different domains, thus confirming the superiority of deep-learning-based SISR.

\section{Conclusions and future work}\label{conclusion}
In this paper we presented a study on how to improve 3D reconstruction starting from low-resolution images through the use of SISR techniques, demonstrating Super-Resolution effectiveness for Multi-View Stereo algorithms based on both PatchMatch and Deep Learning.
Moreover, we have demonstrated the existence of a strong correlation between starting images and 3D models qualities and that an increased amount of input information provided by Super-Resolution is effectively translated in more robust and dense representations in the 3D space by Multi-View Stereo pipelines.
We have shown how, despite the Super-Resolution algorithm chosen, the 3D models obtained result to benefit from the Single-Image Super-Resolution improvement of the input images the more do not have a starting high-resolution.
We believe that this approach can be a great step forward in a wide range of scenarios, and that, given the results obtained, it can be of great help in the world of 3D reconstruction to produce new algorithms or improve those currently in use.
We are also convinced there is ample margin of improvement, such as the creation of ad-hoc Super-Resolution algorithms for this task or the creation of end-to-end architectures mixing Super-Resolution and Multi-View Stereo.



\bibliographystyle{IEEEtranS}
\bibliography{biblio}

\end{document}